\documentclass[sigconf]{acmart}
\usepackage{subcaption}
\usepackage{float}
\usepackage{soul}
\usepackage{url}
\usepackage{epstopdf}
\usepackage{multirow}
\usepackage{multicol}
\usepackage{booktabs}
\usepackage{amsfonts}
\usepackage{booktabs}
\usepackage{bbm}
\usepackage{subcaption}
\usepackage{algorithm}
\usepackage{algorithmic}
\usepackage{eqparbox}

\AtBeginDocument{%
  \providecommand\BibTeX{{%
    \normalfont B\kern-0.5em{\scshape i\kern-0.25em b}\kern-0.8em\TeX}}}

\setcopyright{acmcopyright}
\copyrightyear{2020}
\acmYear{2020} 
\setcopyright{iw3c2w3}
\acmConference[WWW '21]{Proceedings of the Web Conference 2021}{April 19--23, 2021}{Ljubljana, Slovenia} 
\acmBooktitle{Proceedings of the Web Conference 2021 (WWW '21), April 19--23, 2021 ,Ljubljana, Slovenia}
\acmPrice{}
\acmDOI{10.1145/3442381.3450068}
\acmISBN{978-1-4503-8312-7/21/04}

\begin{document}

\title{Extract the Knowledge of Graph Neural Networks and Go Beyond it: An Effective Knowledge Distillation Framework}

\author{
Cheng Yang$^{1,2}$, Jiawei Liu$^{1}$, Chuan Shi$^{1,2}$}
\email{{yangcheng, liu\_jiawei, shichuan}@bupt.edu.cn}
\affiliation{
$^1$School of Computer Science, Beijing University of Posts and Telecommunications\\
$^2$Beijing Key Laboratory of Intelligent Telecommunications Software and Multimedia\\
}
\authornote{Corresponding author.}
\renewcommand{\shortauthors}{Cheng Yang, Jiawei Liu and Chuan Shi}

\renewcommand{\algorithmicrequire}{\textbf{Input:}}
\renewcommand{\algorithmicensure}{\textbf{Output:}}
\begin{abstract}

Semi-supervised learning on graphs is an important problem in the machine learning area. In recent years, state-of-the-art classification methods based on graph neural networks (GNNs) have shown their superiority over traditional ones such as label propagation. However, the sophisticated architectures of these neural models will lead to a complex prediction mechanism, which could not make full use of valuable prior knowledge lying in the data, \textit{e.g.}, structurally correlated nodes tend to have the same class. 
In this paper, we propose a framework based on knowledge distillation to address the above issues. Our framework extracts the knowledge of an arbitrary learned GNN model (teacher model), and injects it into a well-designed student model. The student model is built with two simple prediction mechanisms, \textit{i.e.,} label propagation and feature transformation, which naturally preserves structure-based and feature-based prior knowledge, respectively. In specific, we design the student model as a trainable combination of parameterized label propagation and feature transformation modules. As a result, the learned student can benefit from both prior knowledge and the knowledge in GNN teachers for more effective predictions. Moreover, the learned student model has a more interpretable prediction process than GNNs. We conduct experiments on five public benchmark datasets and employ seven GNN models including GCN, GAT, APPNP, SAGE, SGC, GCNII and GLP as the teacher models. Experimental results show that the learned student model can consistently outperform its corresponding teacher model by $1.4\%\sim 4.7\%$ on average. Code and data are available at \url{https://github.com/BUPT-GAMMA/CPF}
\end{abstract}



\maketitle

\section{Introduction}
Semi-supervised learning on graph-structured data aims at classifying every node in a network given the network structure and a subset of nodes labeled. As a fundamental task in graph analysis~\cite{bhagat2011node}, the classification problem has a wide range of real-world applications such as user profiling~\cite{li2014user}, recommender systems~\cite{tang2009scalable}, text classification~\cite{aggarwal2012survey} and sociological studies~\cite{altenburger2017bias}. Most of these applications have the \textit{homophily phenomenon}~\cite{mcpherson2001birds}, which assumes two linked nodes tend to have similar labels. With the homophily assumption, many traditional methods are developed to propagate labels by random walks~\cite{szummer2002partially,zhu2003semi} or regularize the label differences between neighbors~\cite{joachims2003transductive,zhou2004learning}.

\begin{figure}[t]
    \centering
    \includegraphics[width=0.99\linewidth]{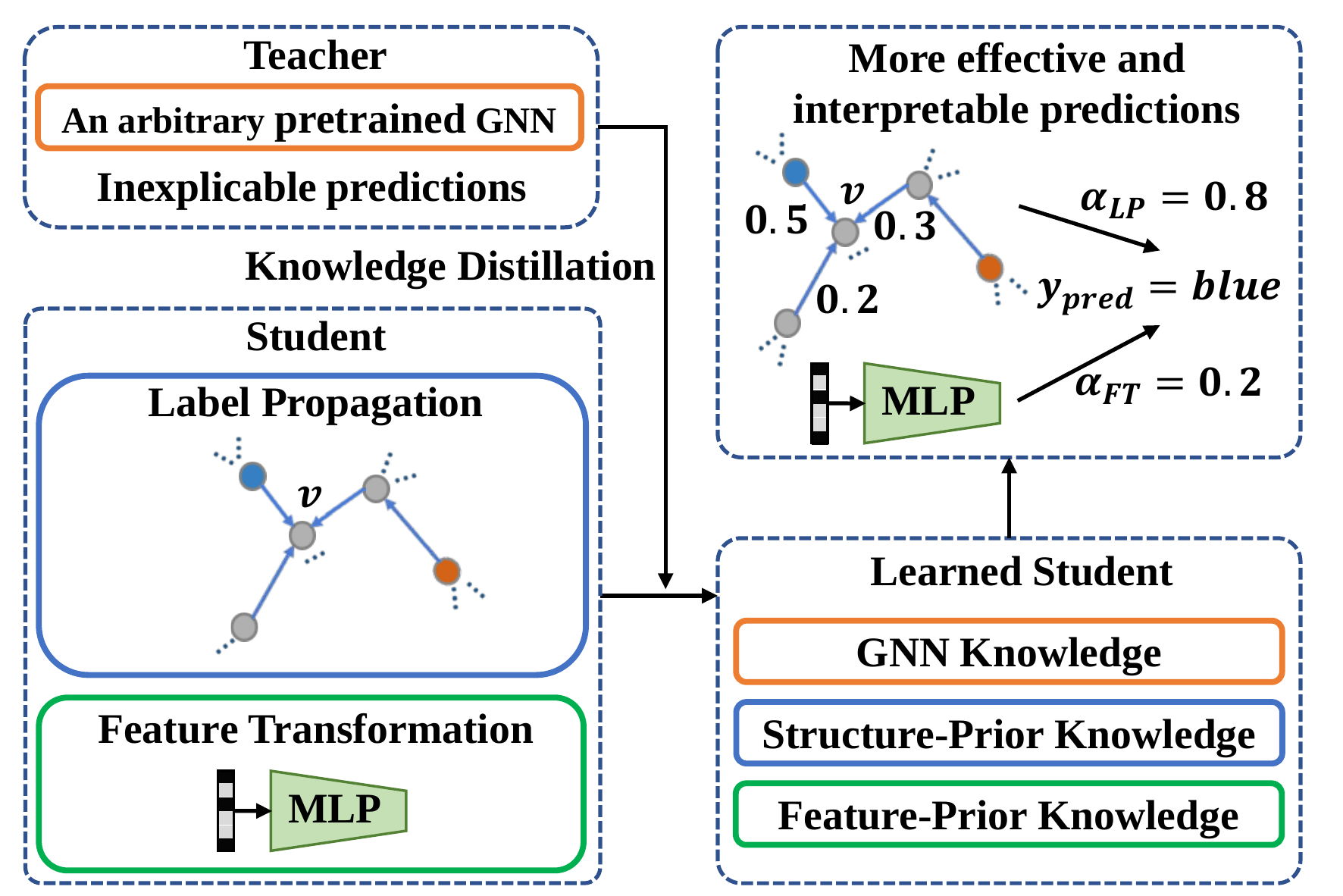}
    \caption{An overview of our knowledge distillation framework. The two simple prediction mechanisms of our student model ensure the full use of structure/feature-based prior knowledge. The knowledge in GNN teachers will be extracted and injected into the student during knowledge distillation. Thus the student can go beyond its corresponding teacher with more effective predictions.}
    \label{fig:intro}
\end{figure}

With the success of deep learning, methods based on graph neural networks (GNNs)~\cite{kipf2016semi,velivckovic2018graph,hamilton2017inductive} have demonstrated their effectiveness in classifying node labels. Most GNN models adopt message passing strategy~\cite{gilmer2017neural}: each node aggregates features from its neighborhood and then a layer-wise projection function with a non-linear activation will be applied to the aggregated information. In this way, GNNs can utilize both graph structure and node feature information in their models.

However, the entanglement of graph topology, node features and projection matrices in GNNs leads to a complicated prediction mechanism and could not take full advantage of prior knowledge lying in the data. For example, the aforementioned homophily assumption adopted in label propagation methods represents \textit{structure-based prior}, and has been shown to be underused~\cite{li2019label,wang2020unifying} in graph convolutional network (GCN)~\cite{kipf2016semi}.

As an evidence, recent studies proposed to incorporate the label propagation mechanism into GCN by adding regularizations~\cite{wang2020unifying} or manipulating graph filters~\cite{li2019label,shi2020masked}. Their experimental results show that GCN can be improved by emphasizing such structure-based prior knowledge. Nevertheless, these methods have three major drawbacks: (1) The main bodies of their models are still GNNs and thus hard to fully utilize the prior knowledge; (2) They are single models rather than frameworks, and thus not compatible with other advanced GNN architectures; (3) They ignored another important prior knowledge, \textit{i.e.}, \textit{feature-based prior}, which means that a node's label is purely determined by its own features.

To address these issues, we propose an effective knowledge distillation framework to inject the knowledge of an arbitrary learned GNN (teacher model) into a well-designed student model. The student model is built with two simple prediction mechanisms, \textit{i.e.}, label propagation and feature transformation, which naturally preserves structure-based and feature-based prior knowledge, respectively. In specific, we design the student model as a trainable combination of parameterized label propagation and feature-based $2$-layer MLP (Multi-layer Perceptron). On the other hand, it has been recognized that the knowledge of a teacher model lies in its soft predictions~\cite{hinton2015distilling}. By simulating the soft labels predicted by a teacher model, our student model is able to further make use of the knowledge in pretrained GNNs. Consequently, the learned student model has a more interpretable prediction process and can utilize both GNN and structure/feature-based priors. An overview of our framework is shown in Fig.~\ref{fig:intro}. 

We conduct experiments on five public benchmark datasets and employ several popular GNN models including GCN~\cite{kipf2016semi}, GAT~\cite{velivckovic2018graph}, SAGE~\cite{hamilton2017inductive}, APPNP~\cite{klicpera2018predict}, SGC~\cite{wu2019simplifying} and a recent deep GCN model GCNII~\cite{chen2020simple} as teacher models. Experimental results show that a student model is able to outperform its corresponding teacher model by $1.4\%\sim 4.7\%$ in terms of classification accuracy. It is worth noting that we also apply our framework on GLP~\cite{li2019label} which unified GCN and label propagation by manipulating graph filters. As a result, we can still gain $1.5\%\sim 2.3\%$ relative improvements, which demonstrates the potential compatibility of our framework. Furthermore, we investigate the interpretability of our student model by probing the learned balance parameters between parameterized label propagation and feature transformation as well as the learned confidence score of each node in label propagation. To conclude, the improvements are consistent and significant with better interpretability.

The contributions of this paper are summarized as follows:

$\bullet$ We propose an effective knowledge distillation framework to extract the knowledge of an arbitrary pretrained GNN model and inject it into a student model for more effective predictions.

$\bullet$  We design the student model as a trainable combination of parameterized label propagation and feature-based $2$-layer MLP. Hence the student model has a more interpretable prediction process and naturally preserves the structure/feature-based priors. Consequently, the learned student model can utilize both GNN and prior knowledge.

$\bullet$ Experimental results on five benchmark datasets with seven GNN teacher models demonstrate the effectiveness of our framework. Extensive studies by probing the learned weights in the student model also illustrate the potential interpretability of our method.
\section{Related Work}
This work is most relevant to graph neural network models and knowledge distillation methods.
\subsection{Graph Neural Networks}
The concept of GNN was proposed~\cite{scarselli2008graph} before 2010 and has become a rising topic since the emergence of GCN~\cite{kipf2016semi}. During the last five years, graph neural network models have achieved promising results in many research areas~\cite{wu2020comprehensive,zhou2018graph}. Now we will briefly introduce some representative GNN methods in this section and employ them as our teacher models in the experiments.

As one of the most influential GNN models, Graph Convolutional Network (GCN)~\cite{kipf2016semi} targeted on semi-supervised learning on graph-structured data through layer-wise propagation of node features. GCN can be interpreted as a variant of convolutional neural networks that operates on graphs. Graph Attention Network (GAT)~\cite{velivckovic2018graph} further employed attention mechanism in the aggregation of neighbors' features. SAGE~\cite{hamilton2017inductive} sampled and aggregated features from a node's local neighborhood and is more space-efficient. Approximate personalized propagation of neural predictions (APPNP)~\cite{klicpera2018predict} studied the relationship between GCN and PageRank, and incorporated a propagation scheme derived from personalized PageRank into graph filters. Simple Graph Convolution (SGC)~\cite{wu2019simplifying} simplified GCN by removing non-linear activations and collapsing weight matrices between layers. Graph Convolutional Network via Initial residual and Identity mapping (GCNII)~\cite{chen2020simple} was a very recent deep GCN model which alleviates the over-smoothing problem. 

Recently, several works show that the performance of GNNs can be further improved by incorporating traditional prediction mechanisms, \textit{i.e.}, label propagation. For example, Generalized Label Propagation (GLP)~\cite{li2019label} modified graph convolutional filters to generate smooth features with graph similarity encoded. UniMP~\cite{shi2020masked} fused feature aggregation and label propagation by a shared message-passing network. GCN-LPA~\cite{wang2020unifying} employed label propagation as regularization to assist GCN for better performances. Note that the label propagation mechanism was built with simple structure-based prior knowledge. Their improvements indicate that such prior knowledge is not fully explored in GNNs. Nevertheless, these advanced models still suffer from several drawbacks as illustrated in the Introduction section.
\subsection{Knowledge Distillation}
Knowledge distillation~\cite{hinton2015distilling} was proposed for model compression where a small light-weight student model is trained to mimic the soft predictions of a pretrained large teacher model. After the distillation, the knowledge in the teacher model will be transferred into the student model. In this way, the student model can reduce time and space complexities without losing prediction qualities. Knowledge distillation is widely used in the computer vision area, \textit{e.g.}, a deep convolutional neural network (CNN) will be compressed into a shallow one to accelerate the inference.

In fact, there are also a few studies combining knowledge distillation with GCN. However, their motivation and model architecture are quite different from ours. Yang et al.~\cite{yang2020distilling} which was proposed in the computer vision area, compressed a deep GCN with large feature maps into a shallow one with fewer parameters using a local structure preserving module. Reliable Data Distillation (RDD)~\cite{zhang2020reliable} trained multiple GCN students with the same architecture and then ensembled them for better performance in a manner similar to BAN~\cite{furlanello2018born}. Graph Markov Neural Networks (GMNN)~\cite{qu2019gmnn} can also be viewed as a knowledge distillation method where two GCNs with different reception sizes learn from each other. Note that both teacher and student models in these works are GCNs.

Compared with them, the goal of our framework is to extract the knowledge of GNNs and go beyond it. Our framework is very flexible and can be applied on an arbitrary GNN model besides GCN. We design a student model with simple prediction mechanisms and thus are able to benefit from both GNN and prior knowledge. As the output of our framework, the student model also has a more interpretable prediction process. In terms of training details, our framework is simpler and requires no ensembling or iterative distillations between teacher and student models for improving classification accuracies.

\begin{figure*}[t]
    \centering
    \includegraphics[width=0.99\linewidth]{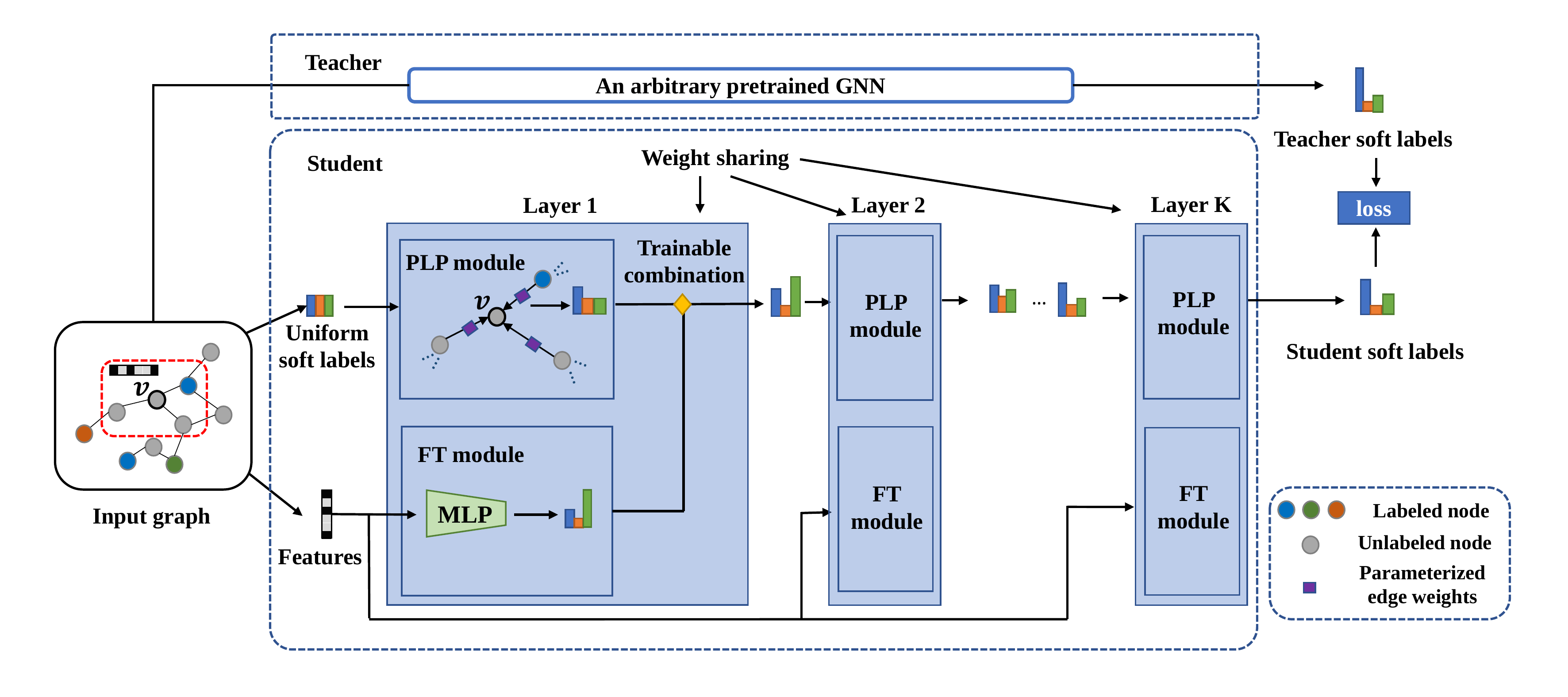}
    \caption{An illustration of the architecture of our proposed student model. Taking the center node $v$ as an example, the student model starts from node $v$'s raw features and a uniform label distribution as soft labels. Then at each layer, the soft label prediction of $v$ will be updated as a trainable combination of Parameterized Label Propagation (PLP) from $v$'s neighbors and Feature Transformation (FT) of $v$'s features. Finally, the distance between the soft label predictions of student and pretrained teacher will be minimized.}
    \label{fig:model}
\end{figure*}

\section{Methodology}
In this section, we will start by formalizing the semi-supervised node classification problem and introducing the notations. Then we will present our knowledge distillation framework to extract the knowledge of GNNs. Afterwards, we will propose the architecture of our student model, which is a trainable combination of parameterized label propagation and feature-based $2$-layer MLP. Finally, we will discuss the potential interpretability of the student model and the computation complexity of our framework.
\subsection{Semi-supervised Node Classification}
We begin by outlining the problem of node classification. Given a connected graph $G=(V,E)$ with a subset of nodes $V_L\subset V$ labeled, where $V$ is the vertex set and $E$ is the edge set, node classification targets on predicting the node labels for every node $v$ in unlabeled node set $V_U=V\setminus V_L$. Each node $v\in V$ has label $y_v\in Y$ where $Y$ is the set of all possible labels. In addition, node features $X\in \mathbb{R}^{|V|\times d}$ are usually available in graph data and can be utilized for better classification accuracy. Each row $X_v\in \mathbb{R}^{d}$ of matrix $X$ denotes a $d$-dimensional feature vector of node $v$.
\subsection{The Knowledge Distillation Framework}
Node classification approaches including GNNs can be summarized as a black box that outputs a classifier $f$ given graph structure $G$, labeled node set $V_L$ and node feature $X$ as inputs. The classifier $f$ will predict the probability $f(v,y)$ that unlabeled node $v\in V_U$ has label $y\in Y$, where $\sum_{y'\in Y} f(v,y’)=1$. For labeled node $v$, we set $f(v,y)=1$ if $v$ is annotated with label $y$ and $f(v,y')=0$ for any other label $y'$. We use $f(v)\in \mathbb{R}^{|Y|}$ to denote the probability distribution over all labels for brevity.

In this paper, the teacher model employed in our framework can be an arbitrary GNN model such as GCN~\cite{kipf2016semi} or GAT~\cite{velivckovic2018graph}. We denote the pretrained classifier in a teacher model as $f_{GNN}$. On the other hand, we use $f_{STU;\Theta}$ to denote the student model parameterized by $\Theta$ and $f_{STU;\Theta}(v)\in \mathbb{R}^{|Y|}$ represents the predicted probability distribution of node $v$ by the student.

In knowledge distillation~\cite{hinton2015distilling}, the student model is trained to mimic the soft label predictions of a pretrained teacher model. As a result, the knowledge lying in the teacher model will be extracted and injected into the learned student. Therefore, the optimization objective which aligns the outputs between the student model and pretrained teacher model can be formulated as
\begin{equation}
    \min_\Theta \sum_{v\in V} distance(f_{GNN}(v),f_{STU;\Theta}(v)),
\label{eq:kd}
\end{equation}
where $distance(\cdot,\cdot)$ measures the distance between two predicted probability distributions. Specifically, we use Euclidean distance in this work\footnote{We also tried to minimize KL-divergence or maximize cross entropy as alternatives. But we find that Euclidean distance performs best and is more numerically stable.}.
\subsection{The Architecture of Student Model}
We hypothesize that a node's label prediction follows two simple mechanisms: (1) label propagation from its neighboring nodes and (2) a transformation from its own features. Therefore, as shown in Fig.~\ref{fig:model}, we design our student model as a combination of these two mechanisms, \textit{i.e.}, a Parameterized Label Propagation (PLP) module and a Feature Transformation (FT) module, which can naturally preserve the structure/feature-based prior knowledge, respectively. After the distillation, the student will benefit from both GNN and prior knowledge with a more interpretable prediction mechanism.

In this subsection, we will first briefly review the conventional label propagation algorithm. Then we will introduce our PLP and FT modules as well as their trainable combinations. 
\subsubsection{Label Propagation}
Label propagation (LP)~\cite{zhu2002learning} is a classical graph-based semi-supervised learning model. This model simply follows the assumption that nodes linked by an edge (or occupying the same manifold) are very likely to share the same label. Based on this hypothesis, labels will propagate from labeled nodes to unlabeled ones for predictions.

Formally, we use $f_{LP}$ to denote the final prediction of LP and $f_{LP}^k$ to denote the prediction of LP after $k$ iterations. In this work, we initialize the prediction of node $v$ as a one-hot label vector if $v$ is a labeled node. Otherwise, we will set a uniform label distribution for each unlabeled node $v$, which indicates that the probabilities of all classes are the same at the beginning. The initialization can be formalized as:
\begin{equation}
f_{LP}^0(v)=
\begin{cases}
(0, ...1, ... 0)\in \mathbb{R}^{|Y|},& \forall v\in V_L\\
(\frac{1}{|Y|}, ... \frac{1}{|Y|}, ... \frac{1}{|Y|})\in \mathbb{R}^{|Y|},& \forall v\in V_U
\end{cases}  
\label{eq:init}
\end{equation}
where $f_{LP}^k(v)$ is the predicted probability distribution of node $v$ at iteration $k$. In the $k+1$-th iteration, LP will update the label predictions of each unlabeled node $v\in V_U$ as follows:
\begin{equation}
f_{LP}^{k+1}(v) = (1-\lambda)\frac{1}{|N_v|}\sum_{u\in N_v} f_{LP}^k(u) + \lambda f_{LP}^k(v),
\label{eq:lp}
\end{equation}
where $N_v$ is the set of node $v$'s neighbors in the graph and $\lambda$ is a hyper-parameter controlling the smoothness of node updates.

Note that LP has no parameters to be trained, and thus can not fit the output of a teacher model through end-to-end training. Therefore, we retrofit LP by introducing more parameters to increase its capacity.
\subsubsection{Parameterized Label Propagation Module}
Now we will introduce our Parameterized Label Propagation (PLP) module by further parameterizing edge weights in LP. As shown in Eq.~\ref{eq:lp}, LP model treats all neighbors of a node equally during the propagation. However, we hypothesize that the importance of different neighbors to a node should be different, which determines the propagation intensities between nodes. To be more specific, we assume that the label predictions of some nodes are more ``confident'' than others: \textit{e.g.}, a node whose predicted label is similar to most of its neighbors. Such nodes will be more likely to propagate their labels to neighbors and keep themselves unchanged.

Formally, we will assign a confidence score $c_v\in \mathbb{R}$ to each node $v$. During the propagation, all node $v$'s neighbors and $v$ itself will compete to propagate their labels to $v$. Following the intuition that a larger confidence score will have a larger edge weight, we rewrite the prediction update function in Eq.~\ref{eq:lp} for $f_{PLP}$ as follows: 
\begin{equation}
f_{PLP}^{k+1}(v) = \sum_{u\in N_v\cup \{v\}}w_{uv}f_{PLP}^{k}(u),
\label{eq:plp}
\end{equation}
where $w_{uv}$ is the edge weight between node $u$ and $v$ computed by the following softmax function:
\begin{equation}
w_{uv} = \frac{exp(c_u)}{\sum_{u'\in N_v\cup \{v\}} exp(c_{u'})}.
\label{eq:edge}
\end{equation}

Similar to LP, $f_{PLP}^{0}(v)$ is initialized as Eq.~\ref{eq:init} and $f_{PLP}^{k}(v)$ remains the one-hot ground truth label vector for every labeled node $v\in V_L$ during the propagation.

Note that we can further parameterize confidence score $c_v$ for inductive setting as an optional choice:
\begin{equation}
c_v = z^T X_v,
\label{eq:ind}
\end{equation}
where $z\in \mathbb{R}^{d}$ is a learnable parameter that projects node $v$'s feature into the confidence score.

\subsubsection{Feature Transformation Module}
Note that PLP module which propagates labels through edges emphasizes the structure-based prior knowledge. Thus we also introduce Feature Transformation (FT) module as a complementary prediction mechanism. The FT module predicts labels by only looking at the raw features of a node. Formally, denoting the prediction of FT module as $f_{FT}$, we apply a $2$-layer MLP\footnote{We find that $2$-layer MLP is necessary for increasing the model capacity of our student, though a single layer logistic regression is more interpretable.} followed by a softmax function to transform the features into soft label predictions:
\begin{equation}
f_{FT}(v)=softmax(MLP(X_v)).
\label{eq:ft}
\end{equation}
\subsubsection{A Trainable Combination} 
Now we will combine the PLP and FT modules as the full model of our student. In detail, we will learn a trainable parameter $\alpha_v\in [0,1]$ for each node $v$ to balance the predictions between PLP and FT. In other words, the prediction from FT module will be incorporated into that from PLP at each propagation step. We name the full student model as Combination of Parameterized label propagation and Feature transformation (CPF) and thus the prediction update function for each unlabeled node $v\in V_U$ in Eq.~\ref{eq:plp} will be rewritten as
\begin{equation}
f_{CPF}^{k+1}(v) = \alpha_{v}\sum_{u\in N_v\cup \{v\}}w_{uv}f_{CPF}^{k}(u) + (1-\alpha_{v}) f_{FT}(v),
\label{eq:full}
\end{equation}
where edge weight $w_{uv}$ and initialization $f_{CPF}^{0}(v)$ are the same with PLP module. Whether parameterizing confidence score $c_v$ as Eq.~\ref{eq:ind} or not will lead to inductive/transductive variants CPF-ind/CPF-tra.

\subsection{The Overall Algorithm and Details}
Assuming that our student model has a total of $K$ layers, the distillation objective in Eq.~\ref{eq:kd} can be detailed as:
\begin{equation}
    \min_\Theta \sum_{v\in V_U} \|f_{GNN}(v)-f_{CPF;\Theta}^K(v)\|_2,
\label{eq:kd2}
\end{equation}
where $\|\cdot\|_2$ is the L2-norm and the parameter set $\Theta$ includes the balancing parameters between PLP and FT $\{\alpha_v, \forall v\in V\}$, confidence parameters in PLP module $\{c_v, \forall v\in V\}$ (or parameter $z$ for inductive setting), and the parameters of MLP in FT module $\Theta_{MLP}$. There is also an important hyper-parameter in the distillation framework: the number of propagation layers $K$. Alg.~\ref{alg:frm} shows the pseudo code of the training process.

We implement our framework based on Deep Graph Library (DGL)~\cite{wang2019deep} and Pytorch~\cite{paszke2019pytorch}, and employ an Adam optimizer~\cite{kingma2014adam} for parameter training. Dropout ~\cite{srivastava2014dropout} is also applied to alleviate overfitting.

\begin{algorithm}
    \caption{The proposed knowledge distillation framework.}
    \label{alg:frm}
    \begin{algorithmic}[1]
        \REQUIRE Graph $G=(V,E)$, labeled node set $V_L\subset V$, unlabeled node set $V_U\subset V$, node features $X$ and pretrained GNN classifier $f_{GNN}$.
        \ENSURE The learned student model $f_{CPF}$.
        \STATE Initialize the label prediction $f_{CPF}^0(v)$ for each node $v$ by Eq.~\ref{eq:init};
        \WHILE{not converge} 
        \IF{inductive setting} \label{alg:while}
        \STATE Compute confidence score $c_v$ for each node $v\in V$ by Eq.~\ref{eq:ind};
        \ENDIF
        \STATE Compute edge weight $w_{uv}$ for each edge $(u,v)\in E$ by Eq.~\ref{eq:edge};
        \FORALL{node $v\in V_U$}
        \STATE Compute the prediction of FT module $f_{FT}(v)$ by Eq.~\ref{eq:ft};
        \FOR{k=1,2\dots K}
        \STATE Update the prediction after $k$ layers $f_{CPF}^k(v)$ by Eq.~\ref{eq:full};
        \ENDFOR
        \ENDFOR
        \STATE Update parameters by optimizing Eq.~\ref{eq:kd2}; \label{alg:endwhile}
        \ENDWHILE 
    \end{algorithmic}
\end{algorithm}

\subsection{Discussions on Interpretability and Complexity}
\label{sec:discuss}
In this subsection, we will discuss the interpretability of the learned student model and the complexity of our algorithm.

After the knowledge distillation, our student model CPF will predict the label of a specific node $v$ as a weighted average between the predictions of label propagation and feature-based MLP. The balance parameter $\alpha_v$ indicates whether structure-based LP or feature-based MLP is more important for node $v$'s prediction. LP mechanism is almost transparent and we can easily find out node $v$ is influenced by which neighbor to what extent at each iteration. On the other hand, the understanding of feature-based MLP can be derived by existing works~\cite{ribeiro2016should} or directly looking at the gradients of different features. Therefore, the learned student model has better interpretability than GNN teachers.

The time complexity of each iteration (line \ref{alg:while} to \ref{alg:endwhile} in Alg.~\ref{alg:frm}) and the space complexity of our algorithm are both $O(|E|+d|V|)$, which is linear to the scale of datasets. In fact, the operations can be easily implemented in matrix form and the training process can be finished in seconds on real-world benchmark datasets with a single GPU device. Therefore, our proposed knowledge distillation framework is very time/space-efficient.

\section{Experiments}
\label{sec:exp}
In this section, we will start by introducing the datasets and teacher models used in our experiments. Then we will detail the experimental settings of teacher models and student variants. Afterwards, we will present quantitative results on evaluating semi-supervised node classification. We also conduct experiments under different numbers of propagation layers and training ratios to illustrate the robustness of our algorithm. Finally, we will present qualitative case studies and visualizations for better understandings of the learned parameters in our student model CPF.
\subsection{Datasets}
\begin{table}[htb]
    \centering
    \caption{Dataset statistics.}
    \resizebox{0.98\columnwidth}{!}
	{
    \begin{tabular}{l|cccc}
    \hline
        Dataset & \# Nodes & \# Edges & \# Features & \# Classes \\
    \hline
       Cora & 2,485 & 5,069 & 1,433 & 7 \\
       Citeseer & 2,110 & 3,668 & 3,703 & 6\\
       Pubmed & 19,717 & 44,324 & 500 & 3\\
       A-Computers & 13,381 & 245,778 & 767 & 10\\
       A-Photo & 7,487 & 119,043 & 745 & 8\\
     \hline
    \end{tabular}}
    \label{tab:data}
\end{table}

\begin{table*}[ht]
\centering
\caption{Classification accuracies with teacher models as GCN~\cite{kipf2016semi} and GAT~\cite{velivckovic2018graph}.}
\label{tab:acc1}
\begin{tabular}{|c|c|c|c|c|c|c||c|c|c|c|c|c|}
\hline
\multirow{2}{*}{Datasets} & Teacher & \multicolumn{4}{c|}{Student variants}               & \multirow{2}{*}{+Impv.} & Teacher & \multicolumn{4}{c|}{Student variants}               & \multirow{2}{*}{+Impv.} \\ \cline{2-6} \cline{8-12}
                          & GCN     & PLP    & FT     & CPF-ind         & CPF-tra         &                         & GAT     & PLP    & FT     & CPF-ind         & CPF-tra         &                         \\ \hline
Cora                      & 0.8244  & 0.7522 & 0.8253 & \textbf{0.8576} & 0.8567          & 4.0\%                   & 0.8389  & 0.7578 & 0.8426 & 0.8576          & \textbf{0.8590} & 2.4\%                   \\ \hline
Citeseer                  & 0.7110  & 0.6602 & 0.7055 & 0.7619          & \textbf{0.7652} & 7.6\%                   & 0.7276  & 0.6624 & 0.7591 & 0.7657          & \textbf{0.7691} & 5.7\%                   \\ \hline
Pubmed                    & 0.7804  & 0.6471 & 0.7964 & 0.8080          & \textbf{0.8104} & 3.8\%                   & 0.7702  & 0.6848 & 0.7896 & 0.8011          & \textbf{0.8040} & 4.4\%                   \\ \hline
A-Computers               & 0.8318  & 0.7584 & 0.8356 & \textbf{0.8443} & \textbf{0.8443} & 1.5\%                   & 0.8107  & 0.7605 & 0.8135 & \textbf{0.8190} & 0.8148          & 1.0\%                   \\ \hline
A-Photo                   & 0.9072  & 0.8499 & 0.9265 & \textbf{0.9317} & 0.9248          & 2.7\%                   & 0.8987  & 0.8496 & 0.9190 & \textbf{0.9221} & 0.9199          & 2.6\%                   \\ \hline
\end{tabular}
\end{table*}

\begin{table*}[ht]
\centering
\caption{Classification accuracies with teacher models as APPNP~\cite{klicpera2018predict} and SAGE~\cite{hamilton2017inductive}.}
\label{tab:acc2}
\begin{tabular}{|c|c|c|c|c|c|c||c|c|c|c|c|c|}
\hline
\multirow{2}{*}{Datasets} & Teacher & \multicolumn{4}{c|}{Student variants}               & \multirow{2}{*}{+Impv.} & Teacher & \multicolumn{4}{c|}{Student variants}               & \multirow{2}{*}{+Impv.} \\ \cline{2-6} \cline{8-12}
                          & APPNP   & PLP    & FT     & CPF-ind         & CPF-tra         &                         & SAGE    & PLP    & FT     & CPF-ind         & CPF-tra         &                         \\ \hline
Cora                      & 0.8398  & 0.7251 & 0.8379 & \textbf{0.8581} & 0.8562          & 2.2\%                   & 0.8178  & 0.7663 & 0.8201 & \textbf{0.8473} & 0.8454          & 3.6\%                   \\ \hline
Citeseer                  & 0.7547  & 0.6812 & 0.7580 & \textbf{0.7646} & 0.7635          & 1.3\%                   & 0.7171  & 0.6641 & 0.7425 & 0.7497          & \textbf{0.7575} & 5.6\%                   \\ \hline
Pubmed                    & 0.7950  & 0.6866 & \textbf{0.8102} & 0.8058          & 0.8081 & 1.6\%                   & 0.7736  & 0.6829 & 0.7717 & 0.7948          & \textbf{0.8062} & 4.2\%                   \\ \hline
A-Computers               & 0.8236  & 0.7516 & 0.8176 & \textbf{0.8279} & 0.8211          & 0.5\%                   & 0.7760  & 0.7590 & 0.7912 & 0.7971          & \textbf{0.8199} & 5.7\%                   \\ \hline
A-Photo                   & 0.9148  & 0.8469 & 0.9241 & \textbf{0.9273} & 0.9272          & 1.4\%                   & 0.8863  & 0.8366 & 0.9153 & \textbf{0.9268} & 0.9248          & 4.6\%                   \\ \hline
\end{tabular}
\end{table*}
We use five public benchmark datasets for experiments and the statistics of the datasets are shown in Table~\ref{tab:data}. As previous works~\cite{shchur2018pitfalls,klicpera2019diffusion,sun2020multi} did, we only consider the largest connected component and regard the edges as undirected. The details about the datasets are as follows: 
\begin{itemize}
    \item Cora~\cite{sen2008collective} is a benchmark citation dataset composed of machine learning papers, where each node represents a document with a sparse bag-of-words feature vector. Edges represent citations between documents, and labels specify the research field of each paper. 
    \item Citeseer~\cite{sen2008collective} is another benchmark citation dataset of computer science publications, holding similar configuration to Cora. Citeseer dataset has the largest number of features among all five datasets used in this paper.
    \item Pubmed~\cite{namata2012query} is also a citation dataset, consisting of articles related to diabetes in the PubMed database. The node features are TF/IDF weighted word frequency, and the label indicates the type of diabetes discussed in this article.
    \item A-Computers and A-Photo~\cite{shchur2018pitfalls} are extracted from Amazon co-purchase graph, where nodes represent products, edges represent whether two products are frequently co-purchased or not, features represent product reviews encoded by bag-of-words, and labels are predefined product categories.
\end{itemize}

Following the experimental settings in previous work~\cite{shchur2018pitfalls}, we randomly sample $20$ nodes from each class as labeled nodes, $30$ nodes for validation and all other nodes for test. 
\subsection{Teacher Models and Settings}
For a thorough comparison, we consider seven GNN models as teacher models in our knowledge distillation framework:
\begin{itemize}
    \item GCN~\cite{kipf2016semi} is a classic semi-supervised model which learns node representations by defining convolution operators on graph-structured data. GCN is sensitive to the number of layers and we employ the most widely-used $2$-layer setting in this work.
    \item GAT~\cite{velivckovic2018graph} improves GCN by incorporating attention mechanism which assigns different weights to each neighbor of a node. We use a $2$-layer GAT with $8$ attention heads as our teacher model.
    \item APPNP~\cite{klicpera2018predict} improves GCN by balancing the preservation of local information and the use of a wide range of neighbor information. We employ $2$ layers and $10$ power iteration steps for APPNP.
    \item SAGE~\cite{hamilton2017inductive} learns node embeddings by sampling and aggregating information from a node’s local neighborhood. We employ the SAGE-GCN variant as a teacher model.
    \item SGC~\cite{wu2019simplifying} reduces the extra complexity of GCN by removing the non-linearity between GCN layers and compressing the weight matrices. Similar to GCN, we also use a $2$-layer setting.
    \item GCNII~\cite{chen2020simple} is a deep model which uses initial residual and identity mapping to avoid oversmoothing of GCN model. Here we use $16$ layers GCNII as a teacher.
    \item GLP~\cite{li2019label} is a label-efficient model which combines label propagation with graph convolution operations by a graph filtering framework. GLP has two model variants: GLP-RNM and GLP-AR, and we use the better one for each dataset as our teacher. 
\end{itemize}

The detailed training settings of teacher models are listed in Appendix A.

\subsection{Student Variants and Experimental Settings}
For each dataset and teacher model, we test the following student variants:
\begin{itemize}
    \item PLP: The student variant with only the Parameterized Label Propagation (PLP) mechanism;
    \item FT: The student variant with only the Feature Transformation (FT) mechanism;
    \item CPF-ind: The full model CPF with inductive setting;
    \item CPF-tra: The full model CPF with transductive setting.
\end{itemize}

We randomly initialize the parameters and employ early stopping with a patience of $50$, \textit{i.e.}, we will stop training if the classification accuracy on validation set does not increase for $50$ epochs. For hyper-parameter tuning, we conduct heuristic search by exploring \# layers $K\in \{5,6,7,8,9,10\}$, hidden size in MLP $d_{MLP}\in\{8,16,32,64\}$, dropout rate $dr\in\{0.2,0.5,0.8\}$, learning rate and weight decay of Adam optimizer $lr\in \{0.001,0.005,0.01\},wd\in\{0.0005,0.001,0.01\}$. 

\begin{table*}[ht]
\centering
\caption{Classification accuracies with teacher models as SGC~\cite{wu2019simplifying} and GCNII~\cite{chen2020simple}. }
\label{tab:acc3}
\begin{tabular}{|c|c|c|c|c|c|c||c|c|c|c|c|c|}
\hline
\multirow{2}{*}{Datasets} & Teacher & \multicolumn{4}{c|}{Student variants}               & \multirow{2}{*}{+Impv.} & Teacher & \multicolumn{4}{c|}{Student variants}               & \multirow{2}{*}{+Impv.} \\ \cline{2-6} \cline{8-12}
                          & SGC     & PLP    & FT     & CPF-ind         & CPF-tra         &                         & GCNII   & PLP    & FT     & CPF-ind         & CPF-tra         &                         \\ \hline
Cora                      & 0.8052  & 0.7513 & 0.8173 & 0.8454          & \textbf{0.8487} & 5.4\%                   & 0.8384  & 0.7382 & 0.8431 & 0.8581          & \textbf{0.8590} & 2.5\%                   \\ \hline
Citeseer                  & 0.7133  & 0.6735 & 0.7331 & 0.7470          & \textbf{0.7530} & 5.6\%                   & 0.7376  & 0.6724 & 0.7564 & \textbf{0.7635} & 0.7569          & 3.5\%                   \\ \hline
Pubmed                    & 0.7892  & 0.6018 & 0.8098 & 0.7972          & \textbf{0.8204} & 4.0\%                   & 0.7971  & 0.6913 & 0.7984 & 0.7928          & \textbf{0.8024} & 0.7\%                   \\ \hline
A-Computers               & 0.8248  & 0.7579 & 0.8391 & 0.8367          & \textbf{0.8407} & 1.9\%                   & 0.8325  & 0.7628 & 0.8411 & \textbf{0.8467} & 0.8447          & 1.7\%                   \\ \hline
A-Photo                   & 0.9063  & 0.8318 & 0.9303 & \textbf{0.9397} & 0.9347          & 3.7\%                   & 0.9230  & 0.8401 & 0.9263 & \textbf{0.9352} & 0.9300          & 1.3\%                   \\ \hline
\end{tabular}
\end{table*}

\begin{table}[htb]
\centering
\caption{Classification accuracies with teacher model as GLP~\cite{li2019label}. }
\label{tab:acc4}
\resizebox{0.98\columnwidth}{!}
	{
\begin{tabular}{|c|c|c|c|c|c|c|}
\hline
\multirow{2}{*}{Datasets} & Teacher & \multicolumn{4}{c|}{Student variants}               & \multirow{2}{*}{+Impv.} \\ \cline{2-6}
                          & GLP     & PLP    & FT     & CPF-ind         & CPF-tra         &                         \\ \hline
Cora                      & 0.8365  & 0.7616 & 0.8314 & \textbf{0.8557} & 0.8539          & 2.3\%                   \\ \hline
Citeseer                  & 0.7536  & 0.6630 & 0.7597 & \textbf{0.7696} & \textbf{0.7696} & 2.1\%                   \\ \hline
Pubmed                    & 0.8088  & 0.6215 & 0.7842 & 0.8133          & \textbf{0.8210} & 1.5\%                   \\ \hline
\end{tabular}}
\end{table}
\subsection{Analysis of Classification Results}
Experimental results on five datasets with seven GNN teachers and four student variants are presented in Table~\ref{tab:acc1}, ~\ref{tab:acc2}, ~\ref{tab:acc3} and ~\ref{tab:acc4}\footnote{We omit the results of GLP on A-Computer/A-Photo because GLP performs much worse than other GNN models on these two datasets in our experiments.}. We have the following observations:
\begin{itemize}
    \item The proposed knowledge distillation framework accompanying with the full architecture of student model CPF-ind and CPF-tra, is able to improve the performance of the corresponding teacher model consistently and significantly. For example, the classification accuracy of GCN on Cora dataset is improved from $0.8244$ to $0.8576$. This is because the knowledge of GNN teachers can be extracted and injected into our student model which also benefits from structure/feature-based prior knowledge introduced by its simple prediction mechanism. This observation demonstrates our motivation and the effectiveness of our framework.
    \item Note that the teacher model Generalized Label Propagation (GLP)~\cite{li2019label} has already incorporated the label propagation mechanism in their graph filters. As shown in Table~\ref{tab:acc4}, we can still gain $1.5\%\sim 2.3\%$ relative improvements by applying our knowledge distillation framework, which demonstrates the potential compatibility of our algorithm. 
    \item Among the four student variants, the full model CPF-ind and CPF-tra always perform best (except APPNP teacher on Pubmed dataset) and give competitive results. Thus both structure-based PLP and feature-based FT modules will contribute to the overall improvements. PLP itself performs worst because PLP which has few parameters to learn has a small model capacity and can not fit the soft predictions of teacher models.
    \item The average relative improvements of the seven teachers\\ GCN/GAT/APPNP/SAGE/SGC/GCNII/GLP are 3.9/3.2/1.4/
    4.7/4.1/1.9/2.0\%, respectively. The improvement over APPNP is the smallest. A possible reason is that APPNP preserves a node's own features during the message passing and thus also utilizes the feature-based prior as our FT module does.
    \item The average relative improvements on the five datasets Cora/\\
    Citeseer/Pubmed/A-Computers/A-Photo are 2.9/4.2/2.7/2.1/ 2.7\%, respectively. Citeseer dataset benefits most from our knowledge distillation framework. A possible reason is that Citeseer has the largest number of features and thus the student model also has more trainable parameters to increase its capacity.
\end{itemize}
\subsection{Analysis of Different Numbers of Propagation Layers}
In this subsection, we will investigate the influence of a key hyper-parameter in the architecture of our student model CPF,  \textit{i.e.}, the number of propagation layers $K$. In fact, popular GNN models such as GCN and GAT are very sensitive to the number of layers. A larger number of layers will cause the over-smoothing issue and significantly harm the model performance. Hence we conduct experiments on Cora dataset for further analysis of this hyper-parameter.
\begin{figure}[htb]
\centering
    \begin{subfigure}{0.42\columnwidth}
		\centering
		\includegraphics[width=\columnwidth]{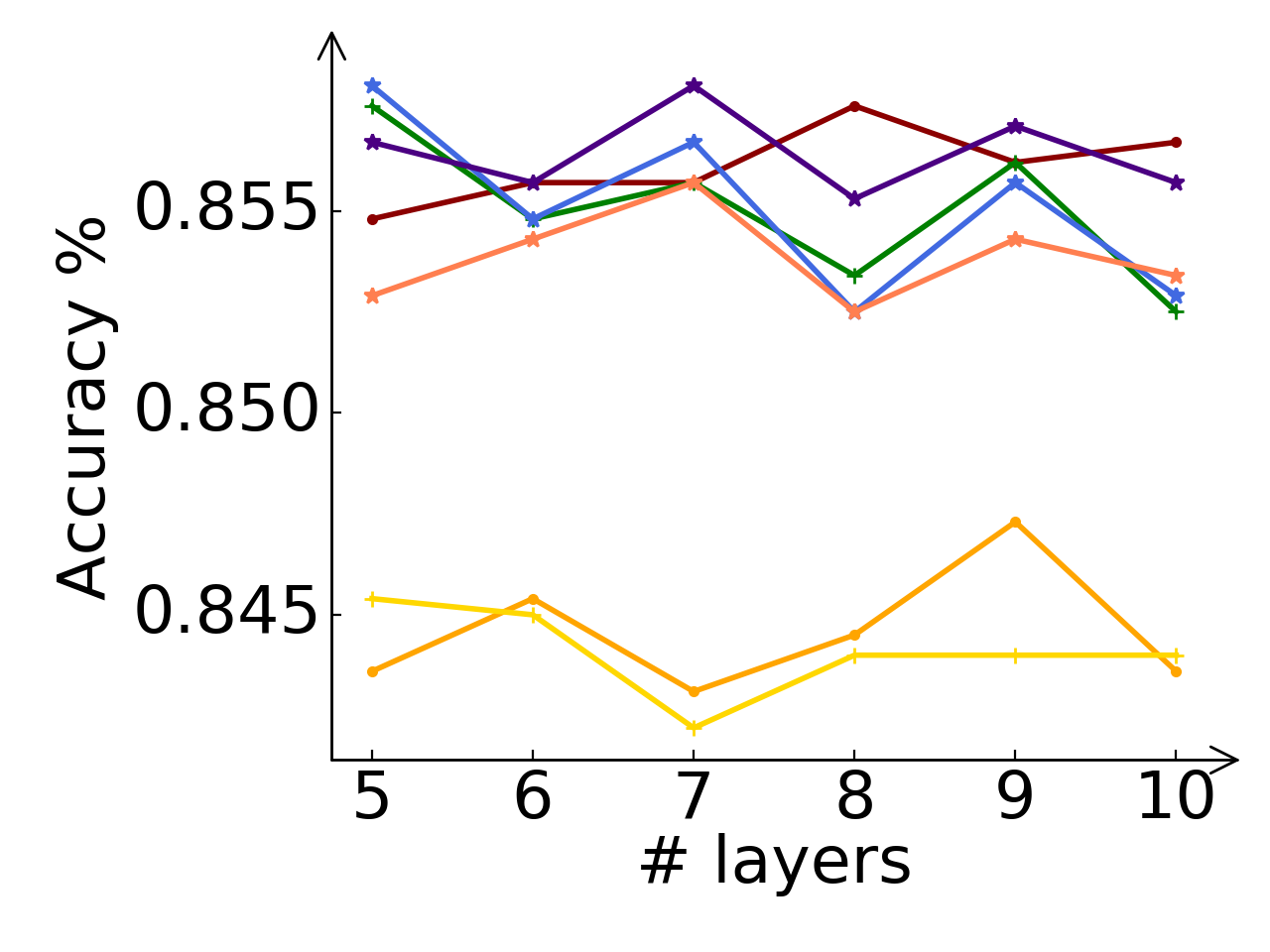}
		\caption{The CPF-ind student.}
	\end{subfigure}
	\begin{subfigure}{0.42\columnwidth}
		\centering
		\includegraphics[width=\columnwidth]{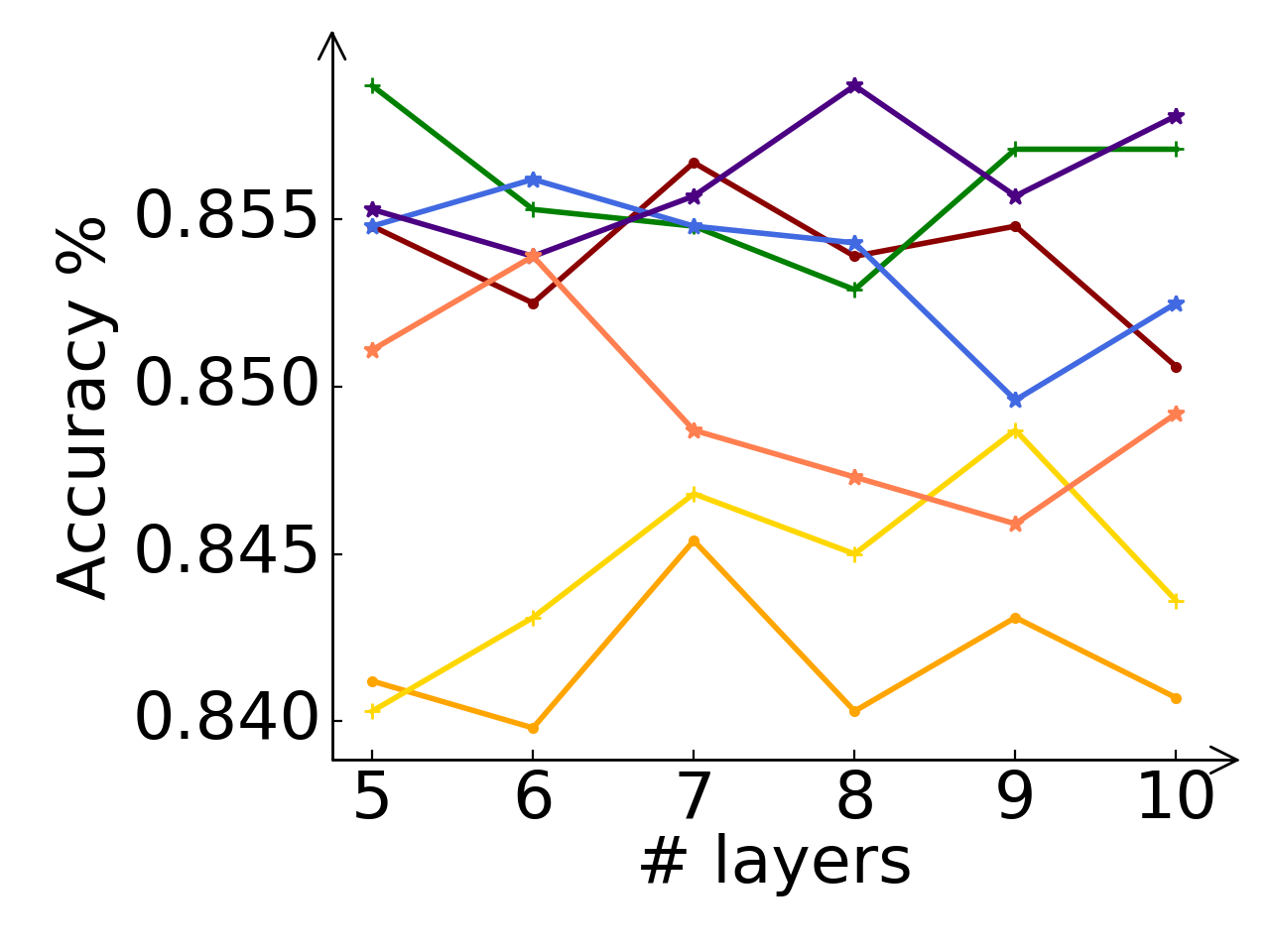}
		\caption{The CPF-tra student.}
	\end{subfigure}
	\begin{subfigure}{0.14\columnwidth}
		\centering
		\includegraphics[width=\columnwidth]{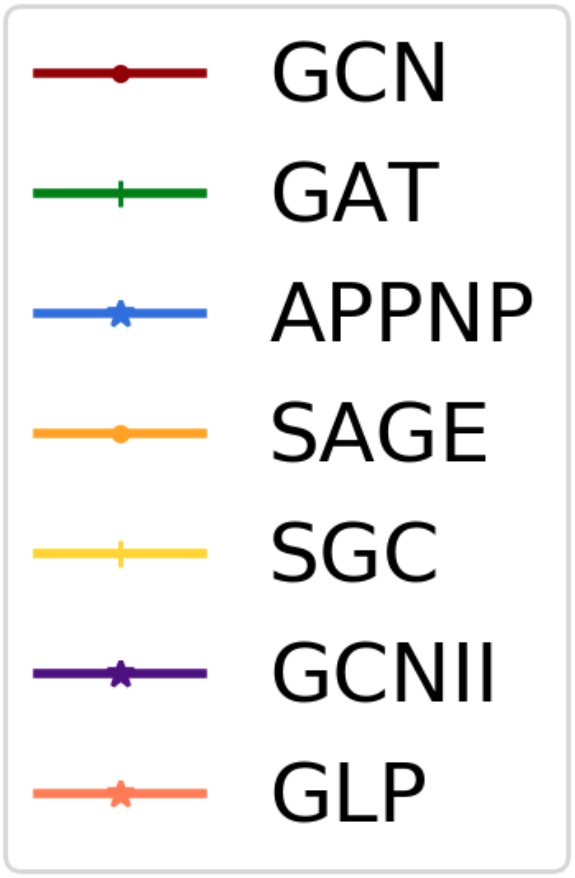}
	\end{subfigure}
\caption{Classification accuracies of CPF-ind and CPF-tra with different numbers of propagation layers on Cora dataset. The legends indicate the teacher model by which a student is guided.}
\label{fig:layer}
\end{figure}

Fig.~\ref{fig:layer} shows the classification results of student CPF-ind and CPF-tra with different numbers of propagation layers $K\in \{5,6,7,8,9,10\}$. We can see that the gaps among different $K$ are relatively small: For each teacher, we compute the gap between the best and worst performed accuracies of its corresponding student and the maximum gaps are $0.56\%$ and $0.84\%$ for CPF-ind and CPF-tra, respectively. Moreover, the accuracy of CPF under the worst choice of $K\in \{5,6,7,8,9,10\}$ has already outperformed the corresponding teacher. Therefore, the gains from our framework are very robust when the number of propagation layers $K$ varies within a reasonable range.

\begin{figure*}[ht]
\begin{minipage}[b]{\linewidth}
    \centering
    \subfloat[][GCN]{\includegraphics[width=.25\linewidth]{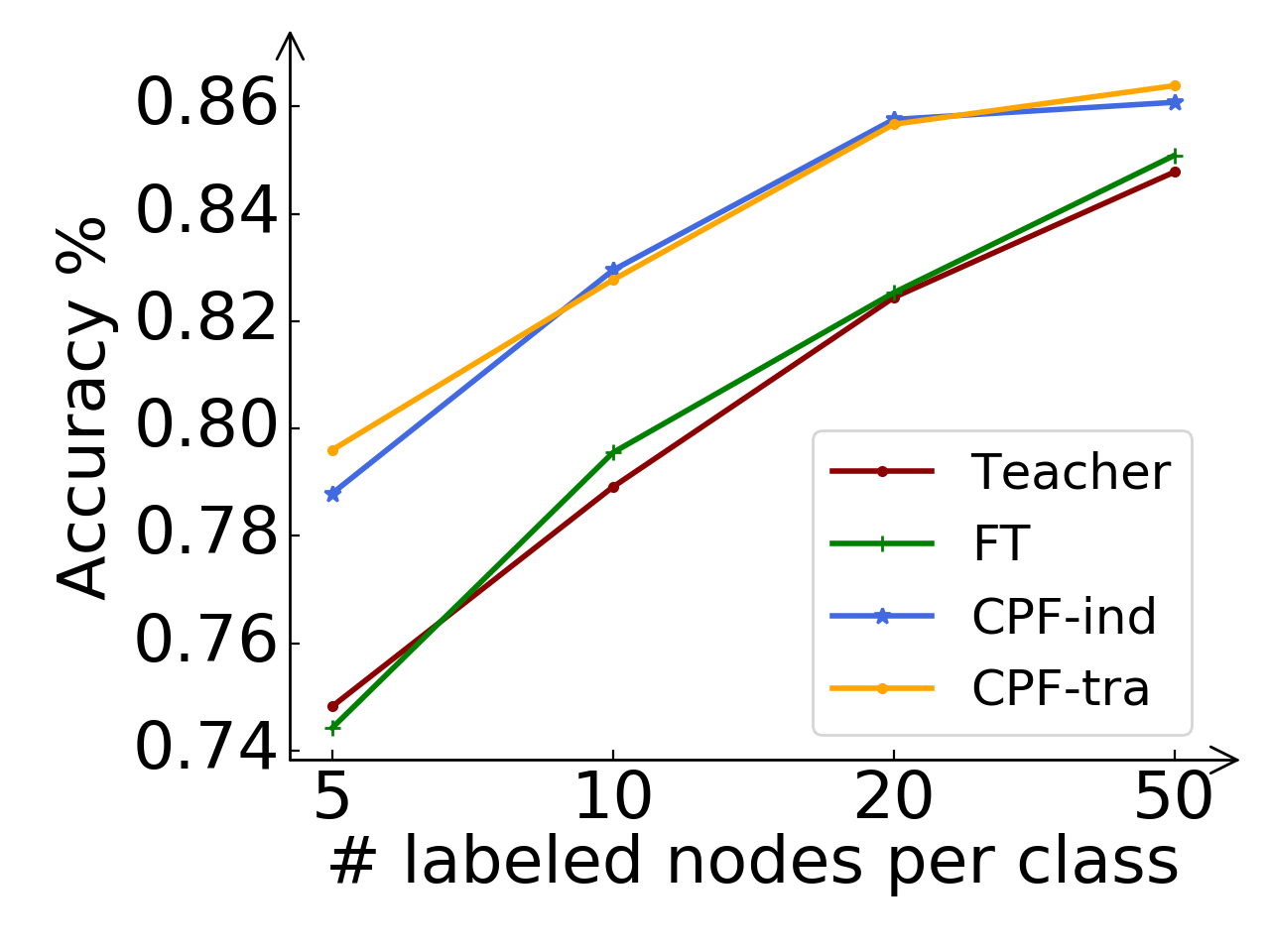}}
    \subfloat[][GAT]{\includegraphics[width=.25\linewidth]{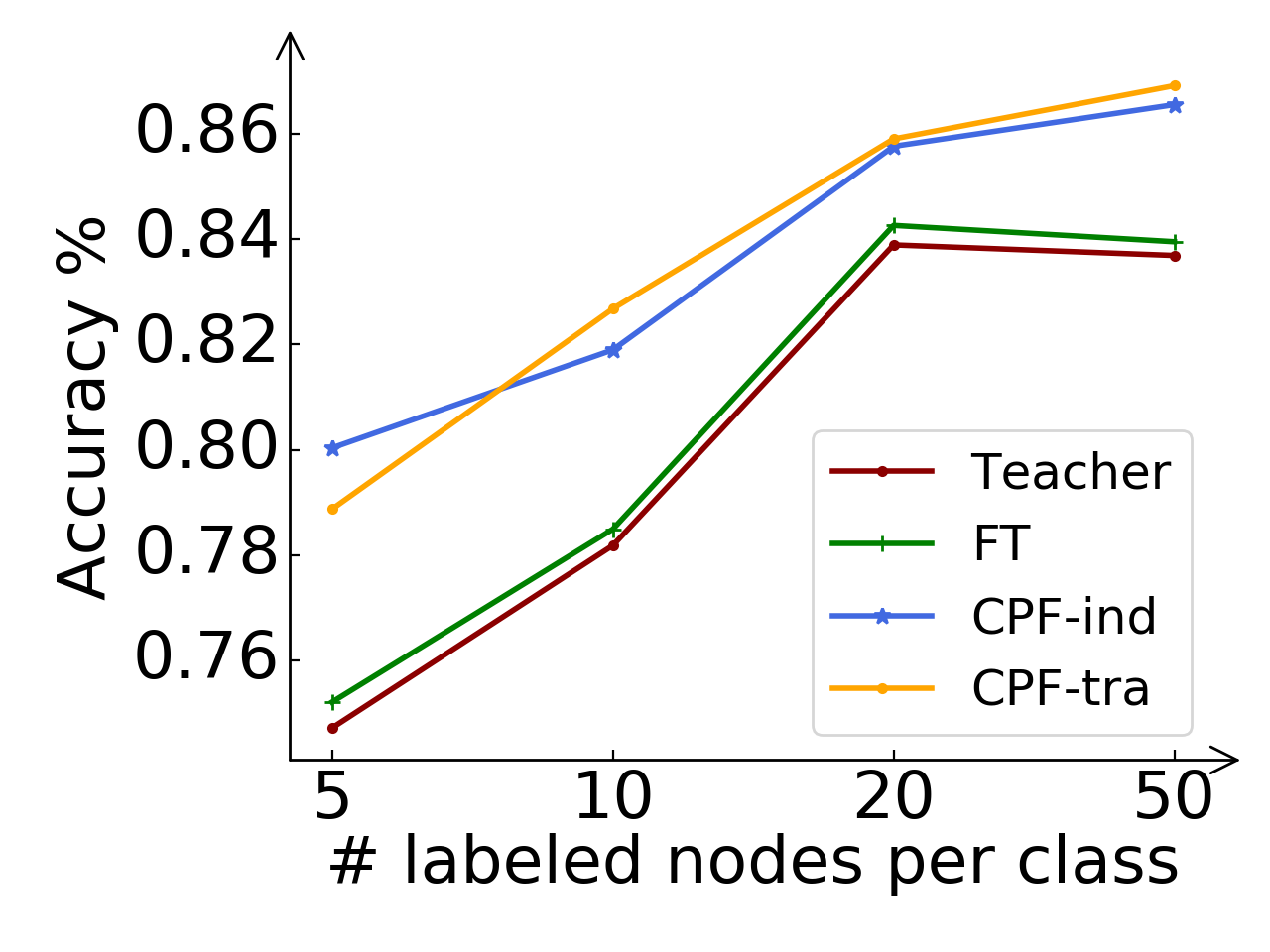}}
    \subfloat[][APPNP]{\includegraphics[width=.25\linewidth]{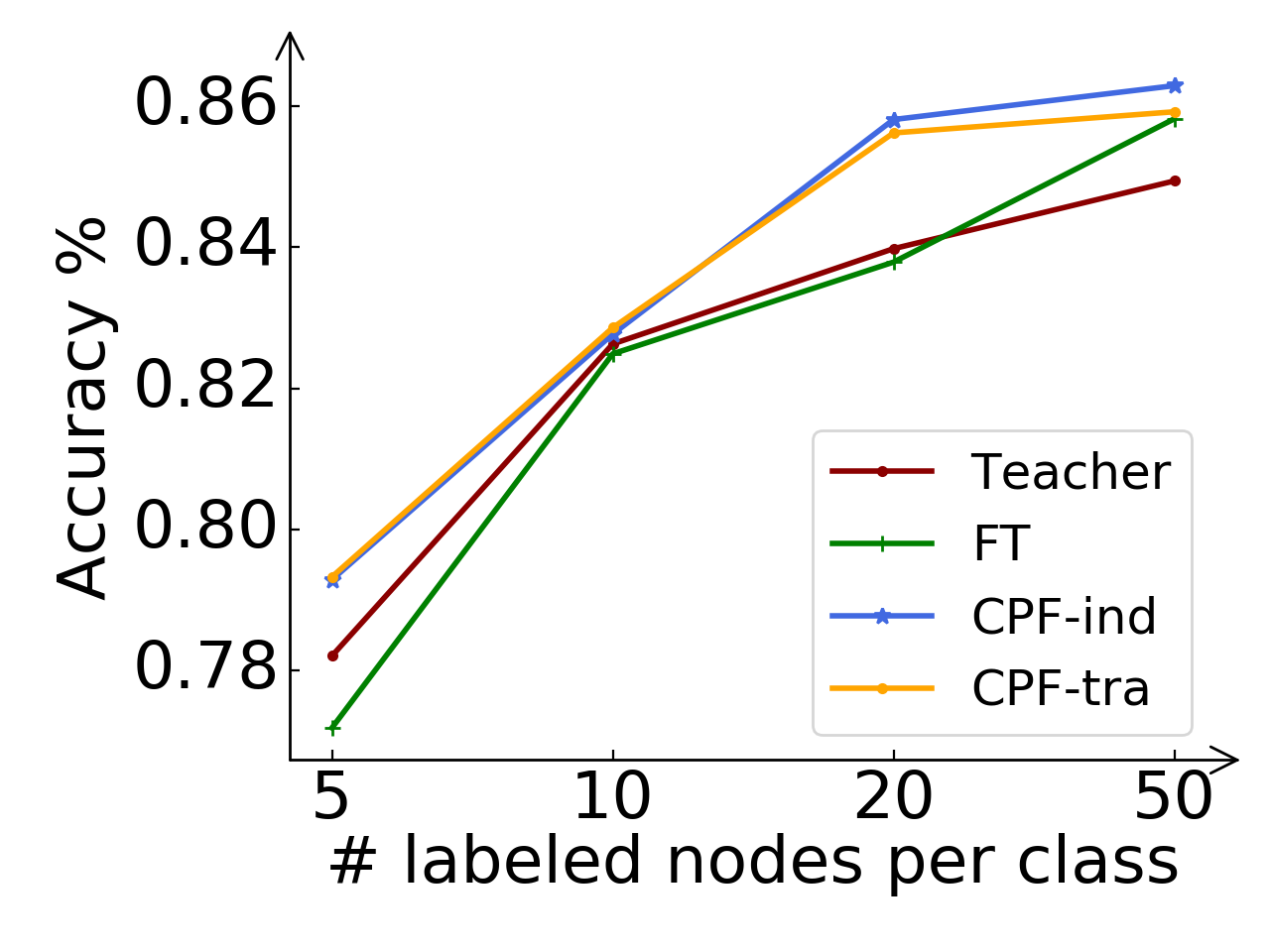}}
    \subfloat[][SAGE]{\includegraphics[width=.25\linewidth]{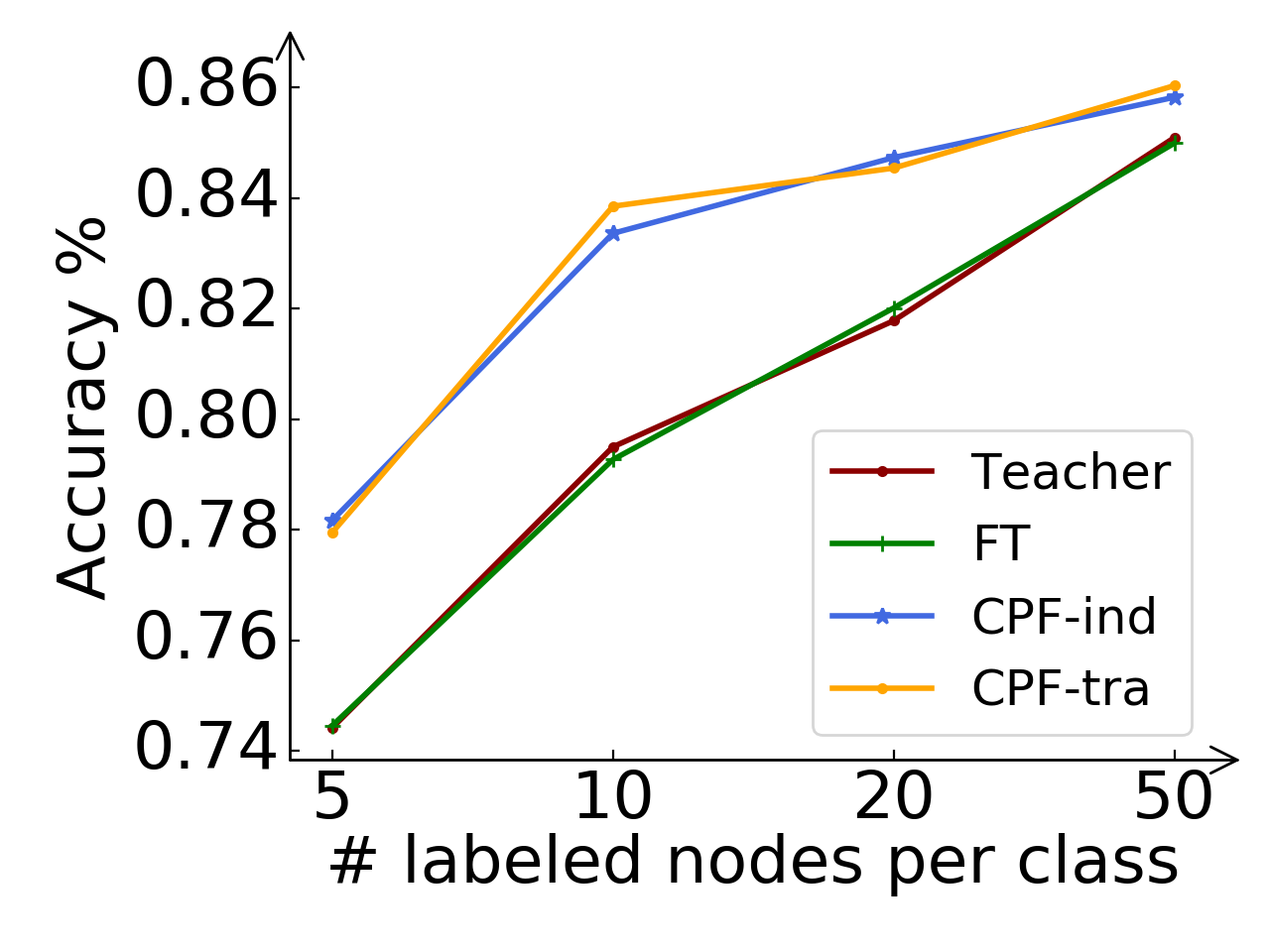}}
\end{minipage}
\medskip
\begin{minipage}[b]{\linewidth}
    \centering
    \subfloat[][SGC]{\includegraphics[width=.25\linewidth]{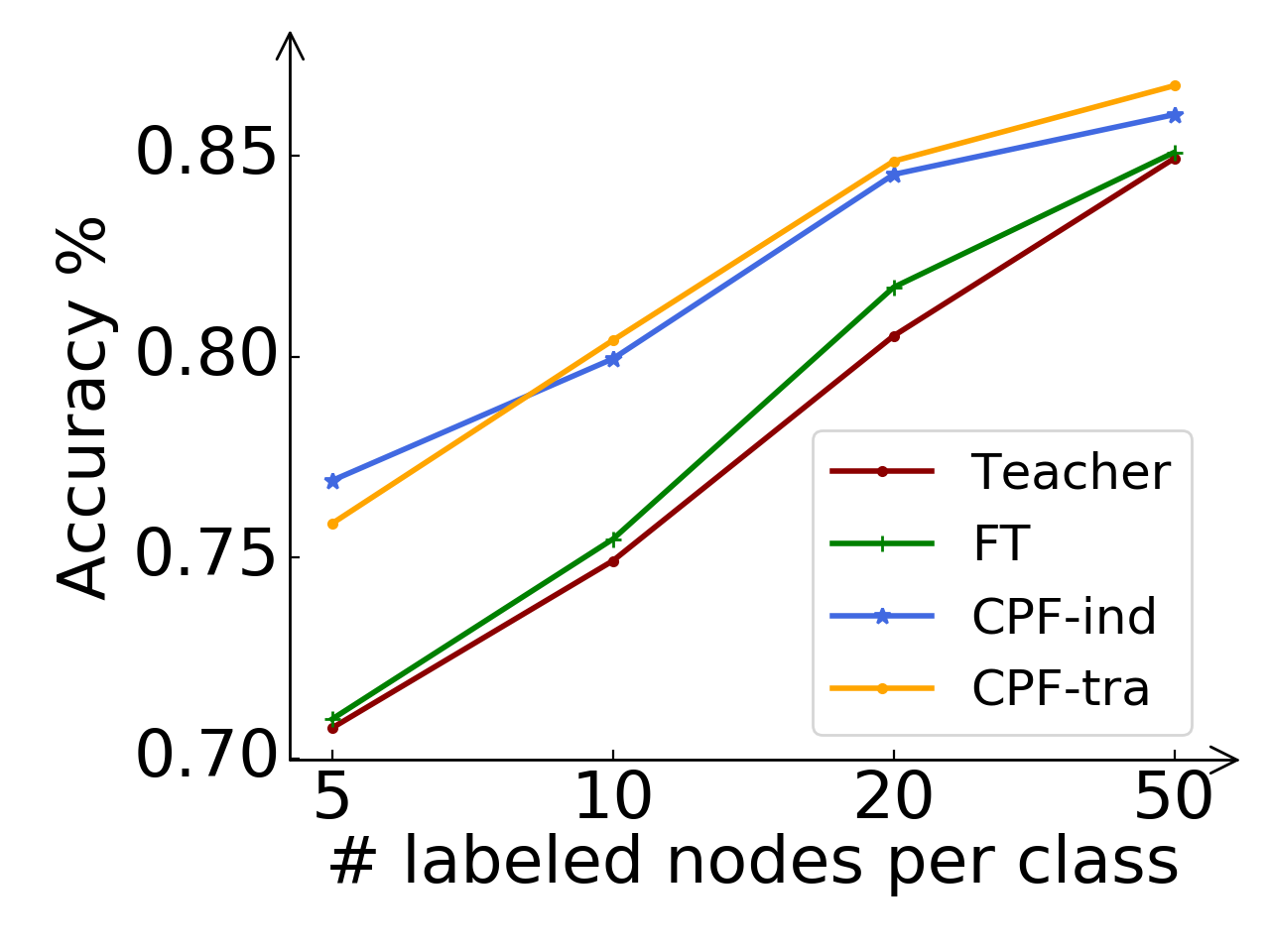}}
    \subfloat[][GCNII]{\includegraphics[width=.25\linewidth]{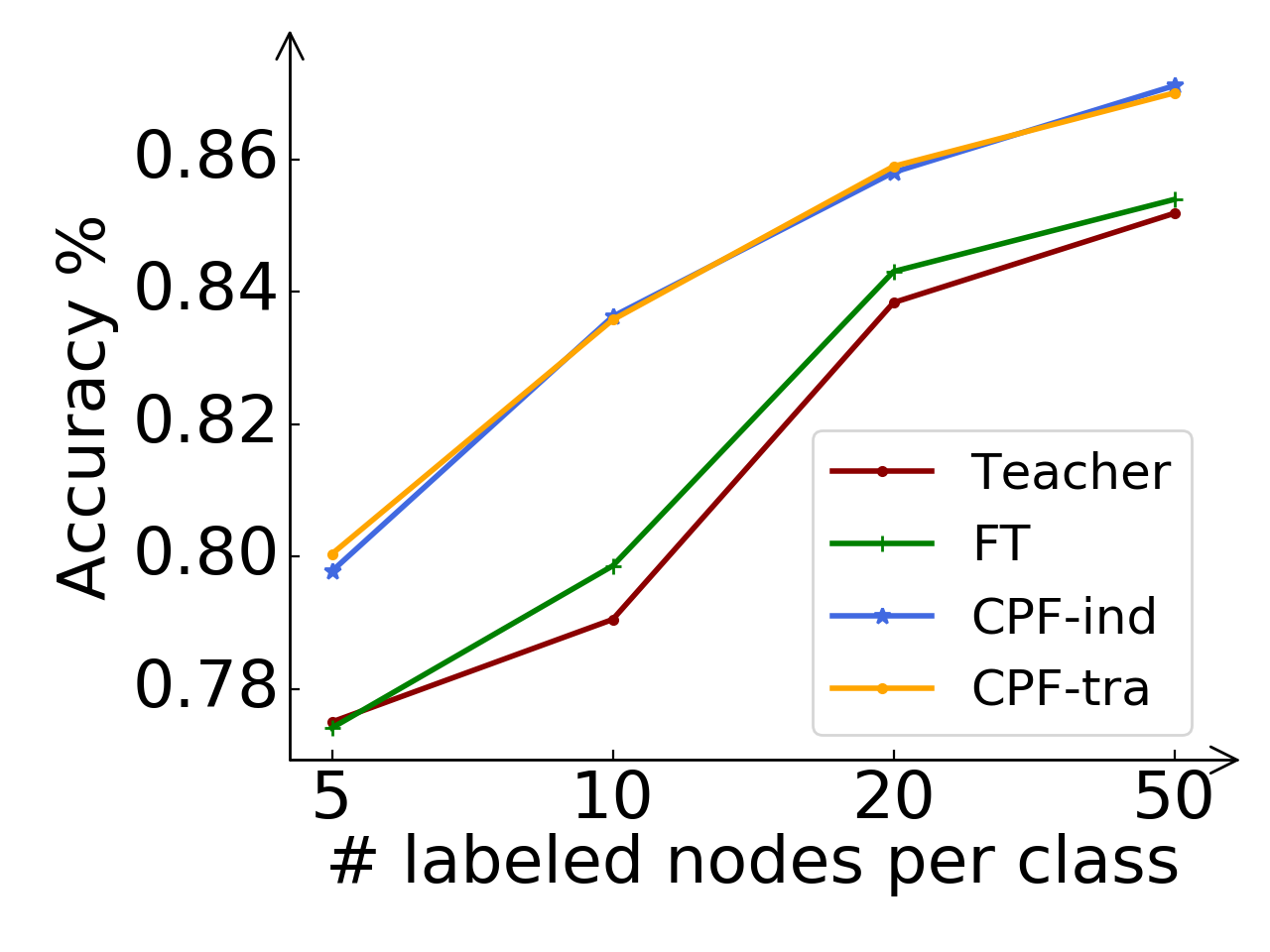}}
    \subfloat[][GLP]{\includegraphics[width=.25\linewidth]{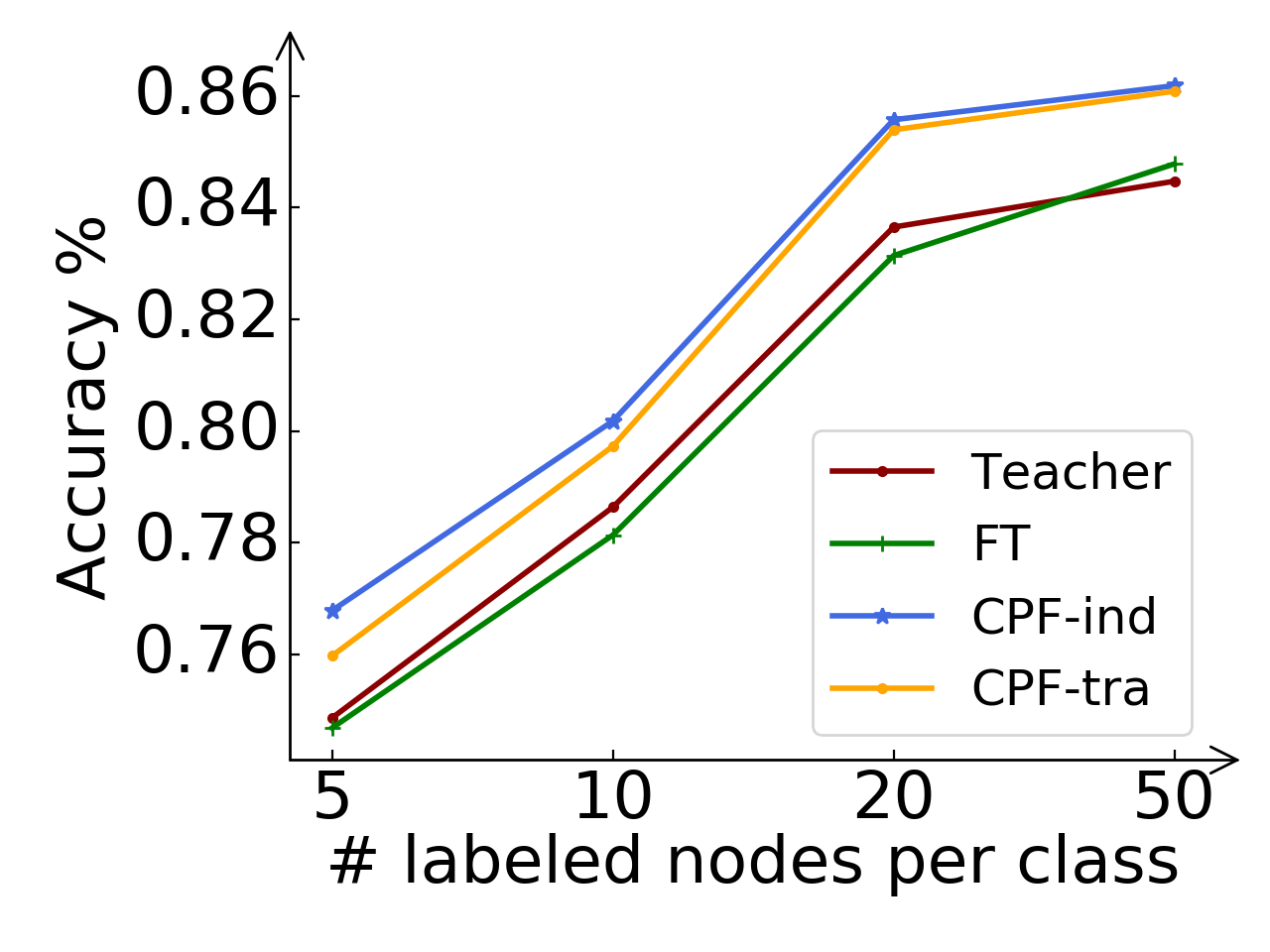}}
\end{minipage} 
\caption{Classification accuracies under different numbers of labeled nodes on Cora dataset. The subcaptions indicate the corresponding teacher models.}
\label{fig:labelnodes}
\end{figure*}

Besides changing the number of propagation layers, another model variant we test is replacing the 2-layer MLP in feature transformation module with a single-layer linear regression, which can also improve the performance with a smaller ratio (the average improvements over the seven teachers are $0.3\%\sim2.3\%$). Linear regression may have better interpretability, but at the cost of weaker performance, which can be seen as a trade-off.
\subsection{Analysis of Different Training Ratios}
To further demonstrate the effectiveness of our framework, we conduct additional experiments under different training ratios. In specific, we take Cora dataset as an example and vary the number of labeled nodes per class from $5$ to $50$. Experimental results are presented in Fig.~\ref{fig:labelnodes}. Note that we omit the results of PLP since its performance is poor and can not be fit into the figures.

We can see that the learned CPF-ind and CPF-tra students consistently outperform the pretrained GNN teachers under different numbers of labeled nodes per class, which illustrates the robustness of our framework. FT module, however, has enough model capacity to overfit the predictions of a teacher but gains no further improvements. Therefore, as a complementary prediction mechanism, the PLP module is also very important in our framework.

Another observation is that the students' improvements over corresponding teacher models are more significant for the few-shot setting, \textit{i.e.}, only $5$ nodes are labeled for each class. As evidence, the relative improvements on classification accuracy are $4.9/4.5/3.2/2.1\%$ on average for $5/10/20/50$ labeled nodes per class. Thus our algorithm also has the ability to handle the few-shot setting which is an important research problem in semi-supervised learning.

\begin{figure*}[ht]
\begin{minipage}[b]{\linewidth}
    \centering
    \subfloat[][Large-$\alpha_v$-GCN (id 1812)]{\includegraphics[width=.25\linewidth]{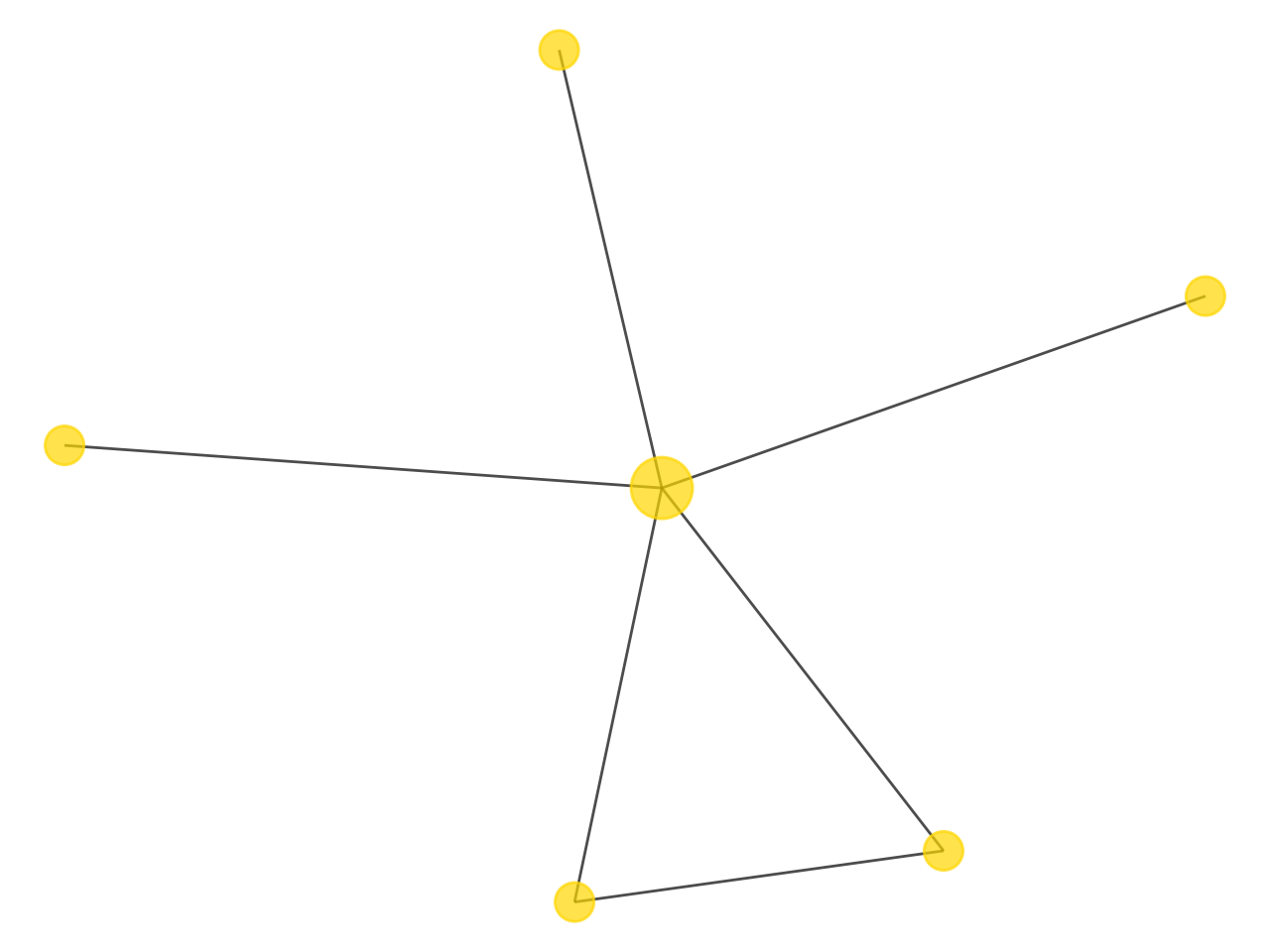}}
    \subfloat[][Large-$\alpha_v$-GAT (id 1720)]{\includegraphics[width=.25\linewidth]{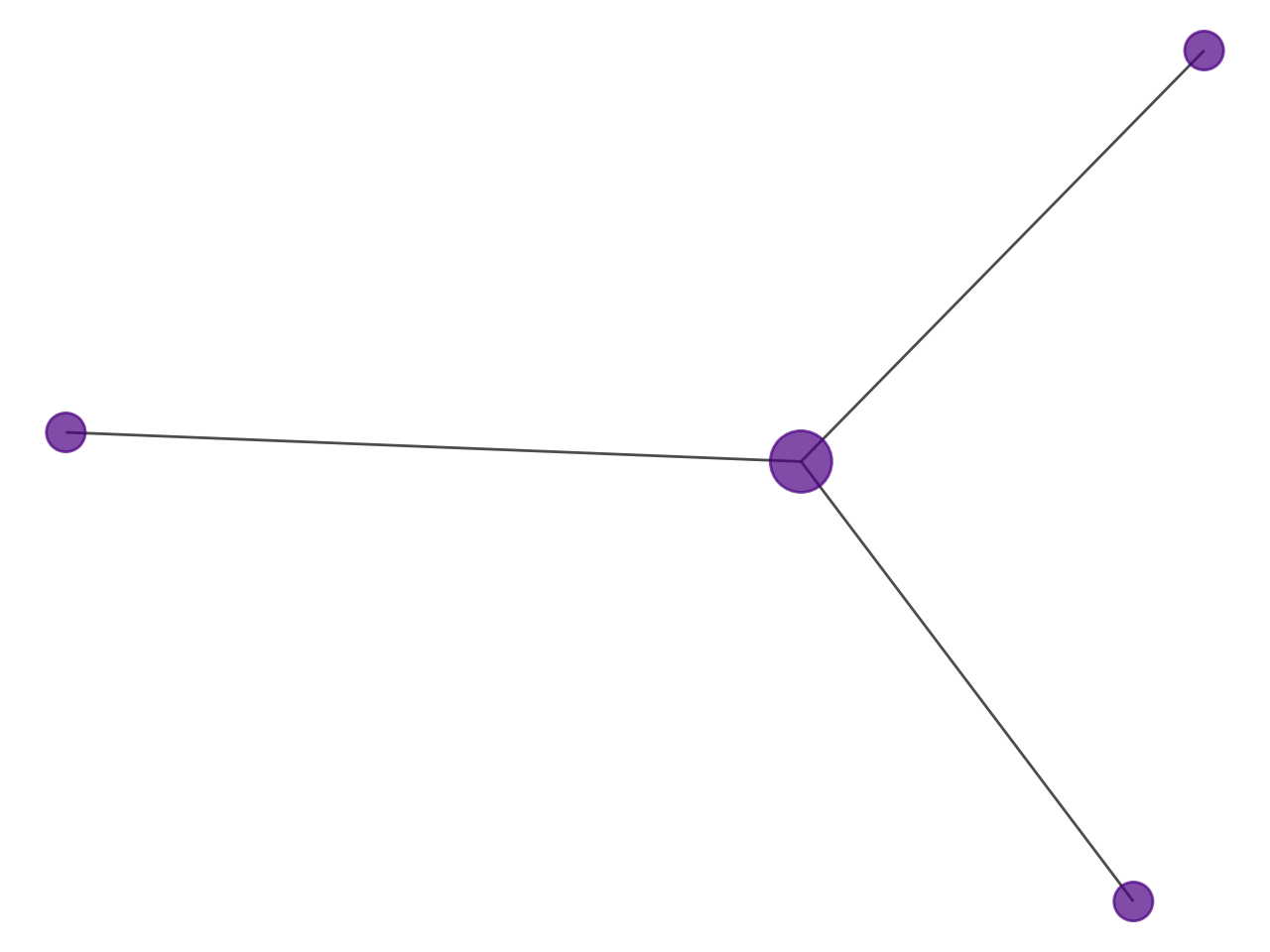}}
    \subfloat[][Small-$\alpha_v$-GCN (id 2381)]{\includegraphics[width=.25\linewidth]{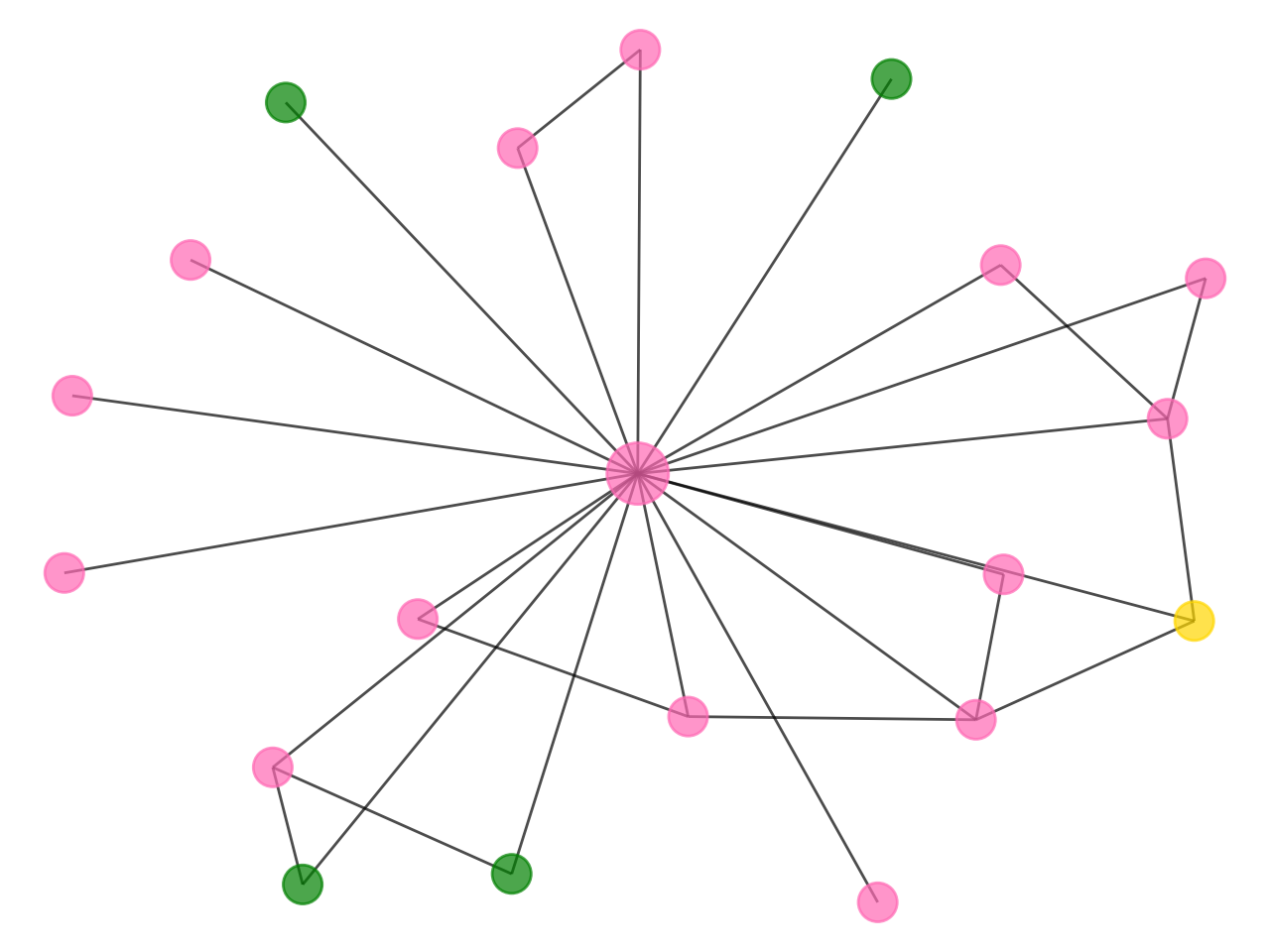}}
    \subfloat[][Small-$\alpha_v$-GAT (id 1298)]{\includegraphics[width=.25\linewidth]{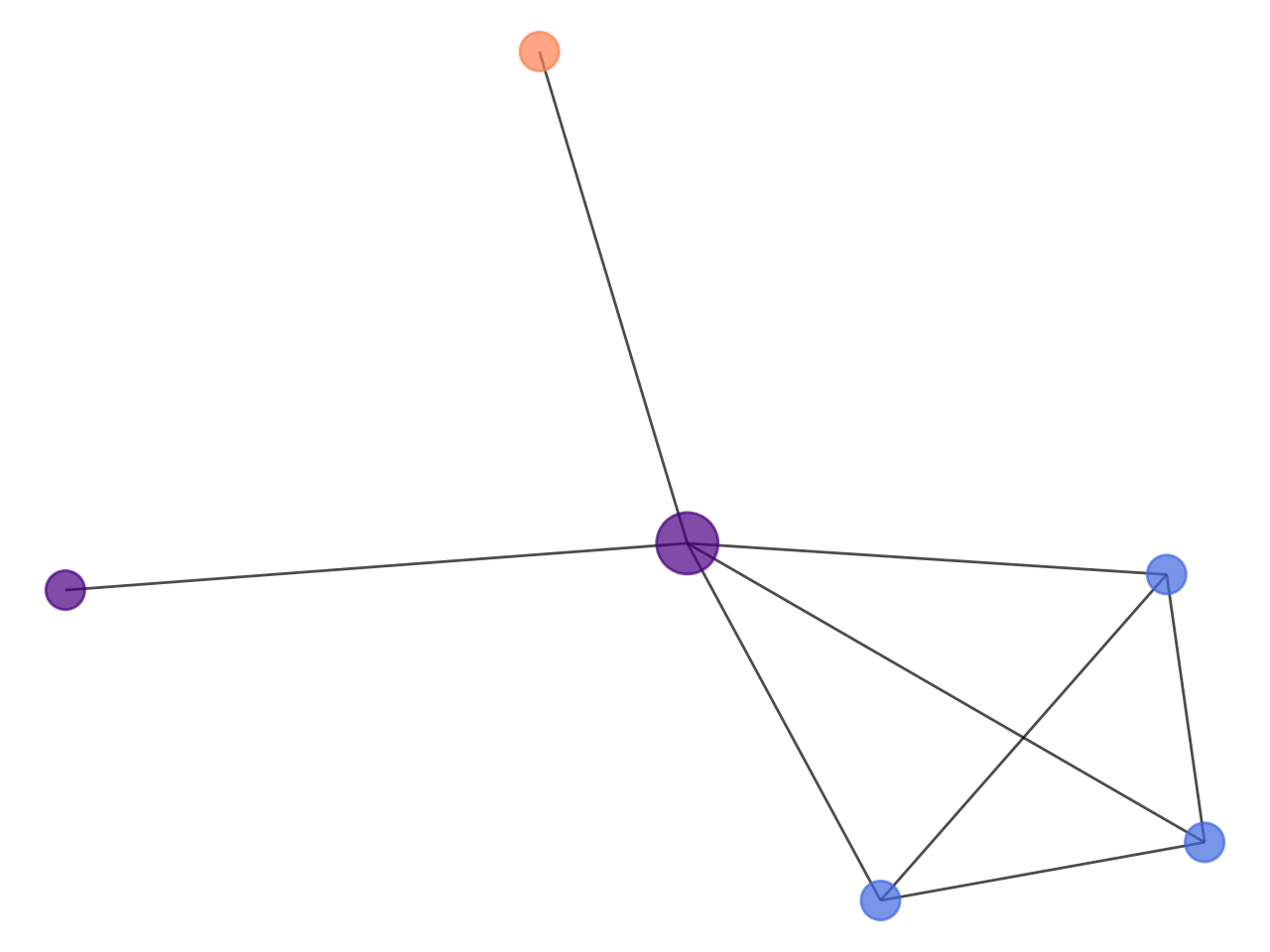}}
\end{minipage}
\caption{Case studies of balance parameter $\alpha_v$ for interpretability analysis. Here the subcaption indicates the node is selected by large/small $\alpha_v$ value with GCN/GAT as teachers.}
\label{fig:alpha}
\end{figure*}
\begin{figure*}[ht]
\begin{minipage}[b]{\linewidth}
    \centering
    \subfloat[][Large-$c_v$-GCN (id 1828)]{\includegraphics[width=.25\linewidth]{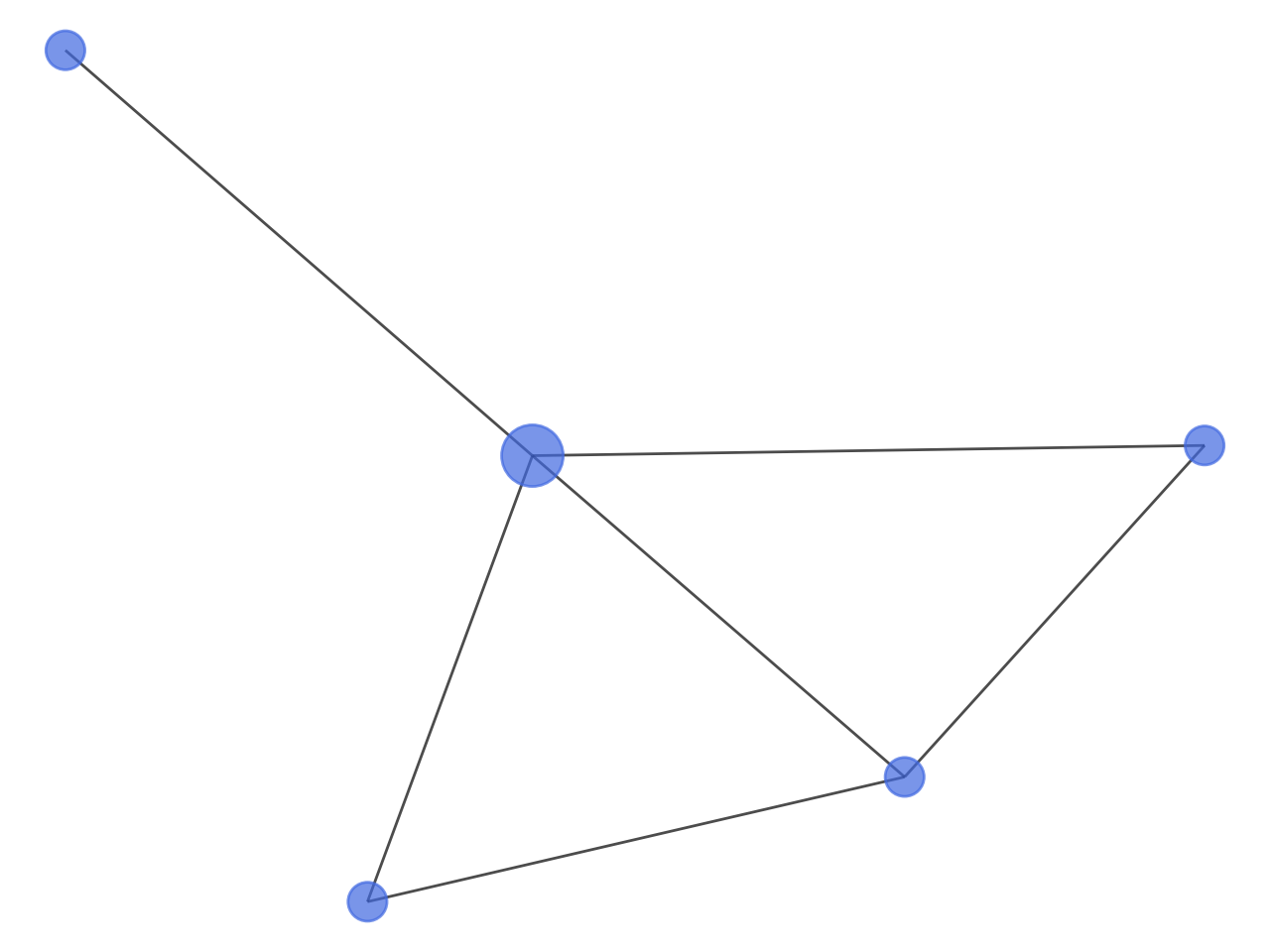}}
    \subfloat[][Large-$c_v$-GAT (id 1450)]{\includegraphics[width=.25\linewidth]{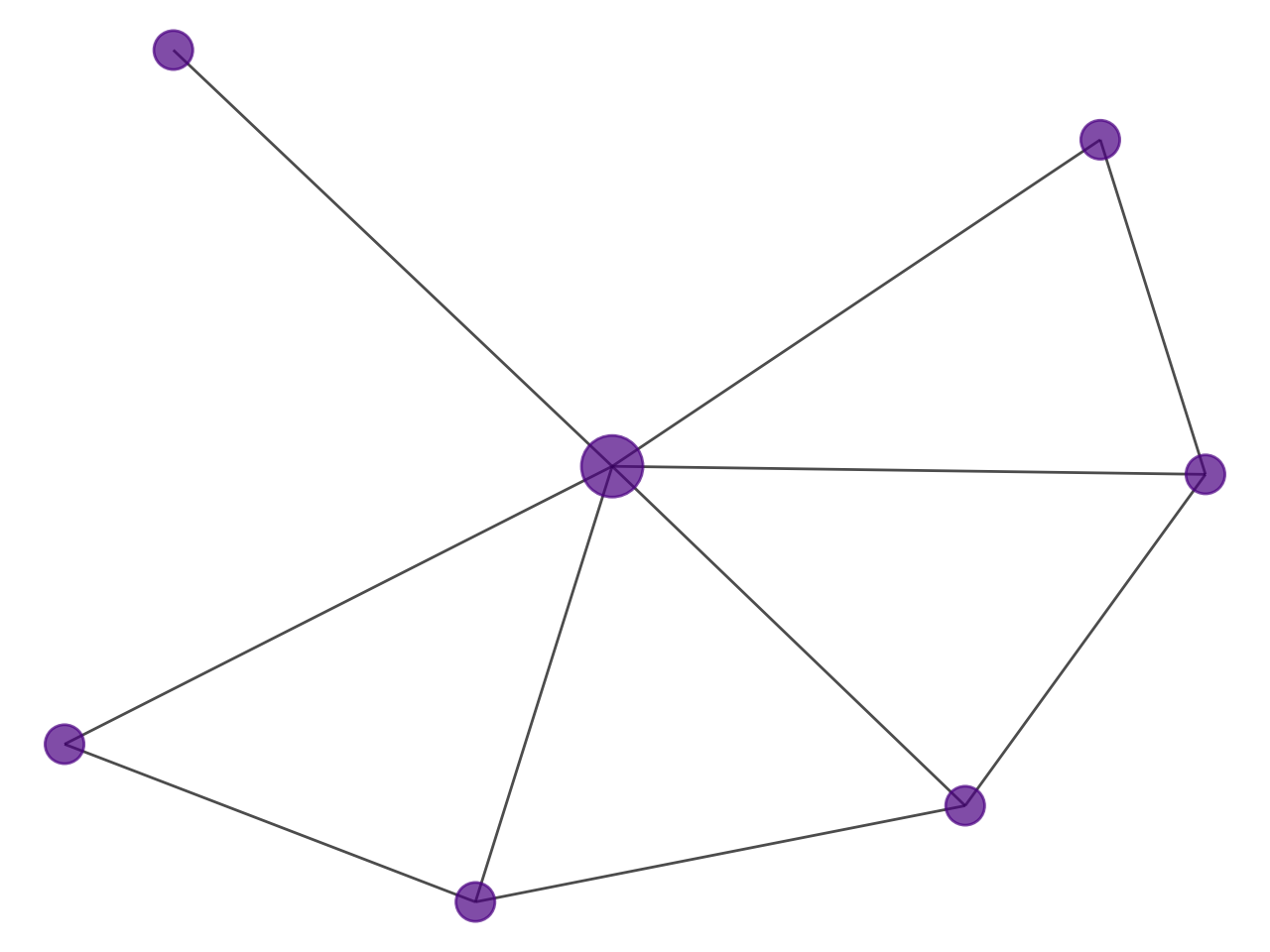}}
    \subfloat[][Small-$c_v$-GCN (id 1238)]{\includegraphics[width=.25\linewidth]{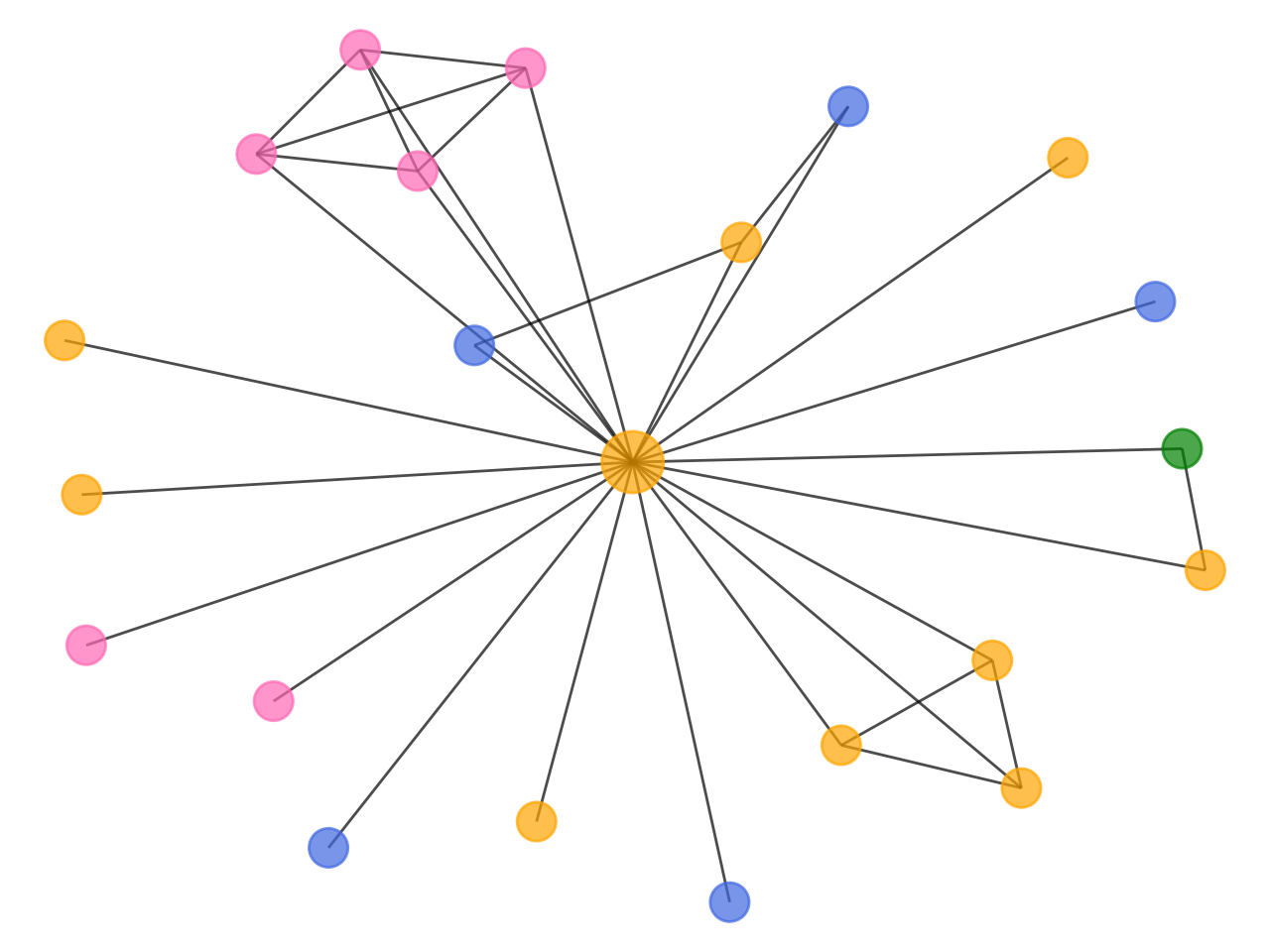}}
    \subfloat[][Small-$c_v$-GAT (id 1160)]{\includegraphics[width=.25\linewidth]{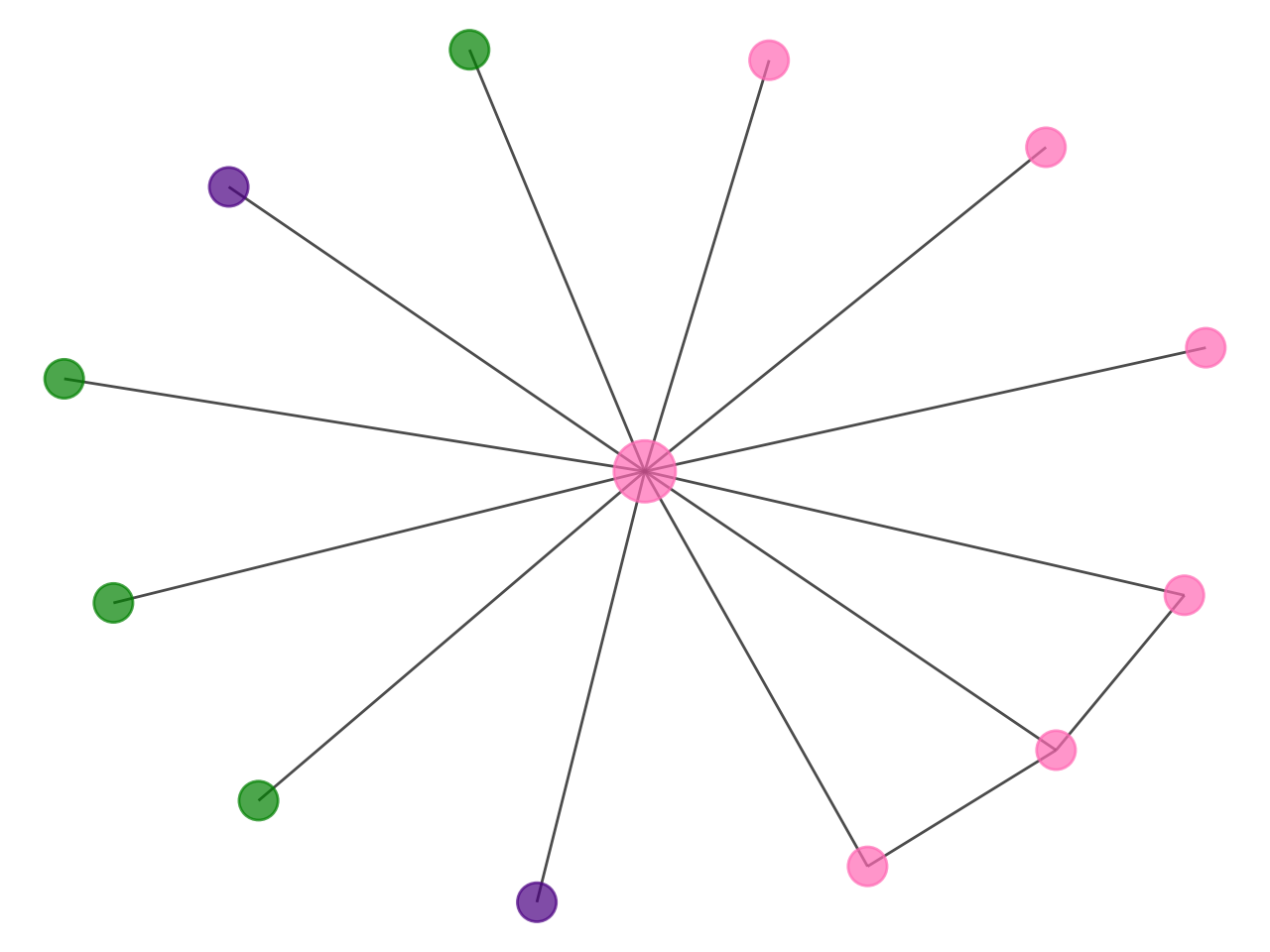}}
\end{minipage} 
\caption{Case studies of confidence score $c_v$ for interpretability analysis. Here the subcaption indicates the node is selected by large/small $c_v$ value with GCN/GAT as teachers.}
\label{fig:confidence}
\end{figure*}
\subsection{Analysis of Interpretability}
Now we will analyze the potential interpretability of the learned student model CPF. Specifically, we will probe into the learned balance parameter $\alpha_v$ between PLP and FT, as well as the confidence score $c_v$ of each node. Our goal is to figure out what kind of nodes has the largest/smallest values of $\alpha_v$ and $c_v$. We use the CPF-ind student guided by GCN or GAT teachers on Cora dataset for illustration in this subsection.

\textbf{Balance parameter $\alpha_v$.} Recall that the balance parameter $\alpha_v$ indicates whether structure-based LP or feature-based MLP contributes more for node $v$'s prediction. As shown in Fig.~\ref{fig:alpha}, we analyze the top-10 nodes with the largest/smallest $\alpha_v$ and select four representative nodes for case study. We plot the 1-hop neighborhood of each node and use different colors to indicate different predicted labels. We find that a node with a larger $\alpha_v$ will be more likely to have the same predicted neighbors. In contrast, a node with a smaller $\alpha_v$ will probably have more neighbors with different predicted labels. This observation matches our intuition that the prediction of a node will be confused if it has many neighbors with various predicted labels and thus can not benefit much from label propagation.

\textbf{Confidence score $c_v$.} On the other hand, a node with a larger confidence score $c_v$ in our student architecture will have larger edge weights to propagate its labels to neighbors and keep itself unchanged. Similarly, as shown in Fig.~\ref{fig:confidence}, we also investigate the top-10 nodes with the largest/smallest confidence score $c_v$ and select four representative nodes for case study. We can see that nodes with high confidences will also have a relatively small degree and the same predicted neighbors. In contrast, nodes with low confidences $c_v$ will have an even more diverse neighborhood than nodes with small $\alpha_v$. Intuitively, a diverse neighborhood of a node will lead to lower confidence to propagate its labels. This finding validates our motivation for modeling node confidences.

\section{Conclusion}
In this paper, we propose an effective knowledge distillation framework which can extract the knowledge of an arbitrary pretrained GNN (teacher model) and inject it into a well-designed student model. The student model CPF is built as a trainable combination of two simple prediction mechanisms: label propagation and feature transformation which emphasize structure-based and feature-based prior knowledge, respectively. After the distillation, the learned student is able to take advantage of both prior and GNN knowledge and thus go beyond the GNN teacher. Experimental results on five benchmark datasets show that our framework can improve the classification accuracies of all seven GNN teacher models consistently and significantly with a more interpretable prediction process. Additional experiments on different numbers of training ratios and propagation layers demonstrate the robustness of our algorithm. We also present case studies to understand the learned balance parameters and confidence scores in our student architecture.

For future work, we will explore the adoption of our framework for other graph-based applications besides semi-supervised node classification. For example, the unsupervised node clustering task would be interesting since the label propagation scheme can not be applied without labels. Another direction is to refine our framework by encouraging the teacher and student models to learn from each other for better performances.

\section{Acknowledgment}
This work is supported by the National Natural Science Foundation of China (No. U20B2045, 62002029, 61772082, 61702296), the Fundamental Research Funds for the
Central Universities 2020RC23, and the National Key Research and Development Program of China (2018YFB1402600).
\bibliographystyle{ACM-Reference-Format}
\bibliography{reference}

\newpage
\appendix
\section{Details for Reproducibility}
In the appendix, we provide more details of experimental settings of teacher models for reproducibility.

The training settings of 5 classical GNNs come from the paper \cite{shchur2018pitfalls}. For the two recent ones (GCNII and GLP), we follow the settings in their original papers. The details are as follows:
\begin{itemize}
    \item GCN~\cite{kipf2016semi}: we use 64 as hidden-layer size, 0.01 as learning rate, 0.8 as dropout probability and 0.001 as learning rate decay.
    \item GAT~\cite{velivckovic2018graph}: we use 64 as hidden-layer size, 0.01 as learning rate, 0.6 as dropout probability, 0.3 as attention dropout probability, and 0.01 as learning rate decay.
    \item APPNP~\cite{klicpera2018predict}: we use 64 as hidden-layer size, 0.01 as learning rate, 0.5 as dropout probability and 0.01 as learning rate decay.
    \item SAGE~\cite{hamilton2017inductive}: we use 128 as hidden-layer size, 0.01 as learning rate, 5 as sample number, 256 as batch size and 0.0005 as learning rate decay.
    \item SGC~\cite{wu2019simplifying}: we use 0.1 as learning rate and 0.001 as learning rate decay.
    \item GCNII~\cite{chen2020simple}: we use 16 as layer number, 64 as hidden-layer size, 0.01 as learning rate, 0.6 as dropout probability, 256 as batch size and 0.1/0.0005 as learning rate decays.
    \item GLP~\cite{li2019label}: we use 16 as hidden-layer size, 0.01 as learning rate, 0.5 as dropout probability, 0.0005 as learning rate decay, k=2 for rnm setting and alpha=10 for ar setting.
\end{itemize}

\end{document}